\documentclass[table, xcdraw]{article}

\usepackage{microtype}
\usepackage{graphicx}
\usepackage{subfigure}
\usepackage{booktabs} 
\usepackage{enumitem}

\usepackage{xcolor}

\usepackage{hyperref}
\usepackage{multirow}

\usepackage[accepted]{icml2025}

\usepackage{amsmath}
\usepackage{amssymb}
\usepackage{mathtools}
\usepackage{amsthm}

\usepackage[capitalize,noabbrev]{cleveref}

\theoremstyle{plain}
\newtheorem{theorem}{Theorem}[section]

\theoremstyle{definition}

\theoremstyle{remark}

\usepackage[textsize=tiny]{todonotes}

\usepackage{multicol}
\usepackage{multirow}
\usepackage{verbatim}

 \icmltitlerunning{~}

\begin{document}
 
\twocolumn[
\icmltitle{Data-Distill-Net: A Data Distillation Approach \\
           Tailored for Reply-based Continual Learning}



\icmlsetsymbol{equal}{*}

\begin{icmlauthorlist}
\icmlauthor{Wenyang Liao}{equal,xjtu}
\icmlauthor{Quanzhiang Wang}{equal,xjtu}
\icmlauthor{Yichen Wu}{hk}
\icmlauthor{Renzhen Wang}{xjtu}
\icmlauthor{Deyu Meng}{xjtu}

\end{icmlauthorlist}

\icmlaffiliation{xjtu}{School of Mathematics and Statistics, Xi'an Jiaotong University, Xi'an, China}
\icmlaffiliation{hk}{Department of Computer Science, City University of Hong Kong}

\icmlcorrespondingauthor{Deyu Meng}{dymeng@mail.xjtu.edu.cn}
\icmlcorrespondingauthor{Renzhen Wang}{rzwang@xjtu.edu.cn}

\icmlkeywords{Machine Learning, ICML}

\vskip 0.3in
]




\printAffiliationsAndNotice{\icmlEqualContribution} 

\begin{abstract}

Replay-based continual learning (CL) methods assume that models trained on a small subset can also effectively minimize the empirical risk of the complete dataset. These methods maintain a memory buffer that stores a sampled subset of data from previous tasks to consolidate past knowledge. However, this assumption is not guaranteed in practice due to the limited capacity of the memory buffer and the heuristic criteria used for buffer data selection.
To address this issue, we propose a new dataset distillation framework tailored for CL, which maintains a learnable memory buffer to distill the global information from the current task data and accumulated knowledge preserved in the previous memory buffer. Moreover, to avoid the computational overhead and overfitting risks associated with parameterizing the entire buffer during distillation, we introduce a lightweight distillation module that can achieve global information distillation solely by generating learnable soft labels for the memory buffer data.
Extensive experiments show that, our method can achieve competitive results and effectively mitigates forgetting across various datasets. The source code will be publicly available.
\end{abstract} 

\section{Introduction}
Continual Learning (CL) \cite{thrun1995lifelong}, also known as Lifelong Learning, is a machine learning paradigm aimed at continually learning new tasks in ever-changing environments while maintaining knowledge from old tasks. This learning approach mimics the human ability to acquire new knowledge without forgetting what has already been mastered. A key challenge of CL is catastrophic forgetting \cite{mccloskey1989catastrophic}, a phenomenon where the model is prone to erasing previously acquired knowledge. The root cause lies in the lack of access to data from prior tasks, causing the model to overwrite parameters containing information relevant to old tasks during the training of new ones, subsequently leading to a substantial decline in performance on earlier tasks.\

To address the issue of catastrophic forgetting in CL, a variety of replay-based works~\cite{ERratcliff1990connectionist,icarl,derpp,clser,RMbang2021rainbow,wu2024meta} have recently emerged. These approaches consolidate previously acquired knowledge by introducing a memory buffer that retains a limited portion of old task data and replays them alongside new task data during training.
For example, the earliest replay-based method, Experience Replay (ER) \cite{ERratcliff1990connectionist}, employs reservoir sampling \cite{Reservoirsamplingvitter1985random} to randomly select samples from old tasks for storage and replay. 
To further refine which samples are retained, numerous coreset selection strategies have been proposed to identify more informative subset for each task. For example, certain heuristic methods concentrate on capturing the most representative samples from each class (e.g., iCaRL~\cite{icarl}, RM \cite{RMbang2021rainbow}, GSS \cite{GSSaljundi2019gradient}, OCS \cite{OCSyoon2021online}) or those lying closer to the decision boundary (e.g., Rwalk \cite{chaudhry2018riemannian}, MIR \cite{aljundi2019online}, ASER \cite{shim2021online}).

All these methods are fundamentally based on the assumption that training on a small coreset is sufficient to minimize the empirical risk on the entire original dataset. However, due to capacity constraints, a limited memory buffer may fail to accurately approximate the stationary distribution of all previously encountered tasks, particularly when exemplars are selected using heuristic strategies \cite{wang2022improving}. Consequently, a substantial discrepancy often arises between the distribution of the stored data and that of the complete dataset from older tasks. This naturally raises the question: \textit{how can a limited buffer be used to better approximate the underlying assumption of replay-based methods?}

Recent research \cite{liu2020mnemonics,LoDM} suggests that rather than merely selecting a subset, generating a small dataset that preserves performance on the original dataset could offer a promising alternative. This learning paradigm, often referred to as dataset distillation or dataset condensation \cite{DDwang2018dataset}, aims to compress the information from a large dataset into a compact synthetic dataset that achieves results on par with the original. In CL scenarios, since only the task $\mathcal T^n$ is accessible during learning task $n$, recent distillation-based methods \cite{liu2020mnemonics,LoDM} focus on compressing $\mathcal T^n$ into a small subset $\mathcal E^n$, which is then combined with the existing memory buffer $\mathcal M^{n-1}=\cup_{k=1}^{n-1} \mathcal E^k$ to form a new buffer, \textit{i.e.}, $\mathcal M^{n}=\mathcal M^{n-1} \cup \mathcal E^n$. Upon analyzing the evolution of the memory buffer, we observe that, up to task timestamp $n$, the resulting $\mathcal M^n$ ensures only intra-task information condensation, which neglects the relationships between different tasks. In other words, while $\mathcal M^n$ can effectively preserve class discrimination within each task $\mathcal{T}^k$ (for $k\in [n]$) \footnote{This is achieved by approximating the empirical risk between $\mathcal{E}^k$ and $\mathcal{T}^k$ during the distillation process.}, it fails to maintain class discrimination across all $n$ tasks. Consequently, this limitation prevents $\mathcal M^n$ from meeting the practical requirements of replay-based methods, which assume on training with \(\mathcal{M}^n\) to adequately minimize the empirical risk over the entire dataset \(\mathcal{T}^{[1:n]}\).

To address this challenge, this paper proposes a simple yet efficient dataset distillation framework tailored for replay-based CL methods. The framework aims to construct a memory buffer distilled not only from the current task data but also from the accumulated knowledge preserved in the previous memory buffer. Unlike current methods \cite{liu2020mnemonics,LoDM}, our method treats the entire memory buffer as differentiable parameters at each task timestamp, enabling it to capture global information from cross-task data. To mitigate the computational overhead and overfitting risk introduced by parameterizing the entire memory buffer, we further propose a novel strategy in which only the labels of the memory buffer data are treated as learnable parameters, with these parameters generated by a unified hyper-network, Data-Distill-Net (or abbreviated as DDN). We theoretically demonstrate the feasibility of distilling global information from cross-task data via the hyper-network and empirically validate the effectiveness of our approach through extensive experiments across various settings and benchmark datasets. Our main contributions are summarized as follows:

\begin{itemize}[leftmargin=4mm, itemsep=-2mm, topsep=-2 mm]
    \item We propose a new dataset distillation framework tailored for CL, which maintains a learnable memory buffer capable of distilling global information from the current task data and accumulated knowledge preserved in the previous memory buffer, thereby mitigating forgetting.
    \item Rather than parameterizing the entire memory buffer, we further propose a light-weight distillation module, Data-Distill-Net, to generate learnable soft labels for memory buffer data, which efficiently mitigates the computational overhead and overfitting risk associated with our proposed distillation framework.
    \item We evaluate the proposed method on four replay-based baselines under both online and offline settings. Experimental results show that our approach effectively consolidates prior knowledge and seamlessly integrates as a plugin to enhance various replay-based baselines.
\end{itemize}

\section{Related Work}
\subsection{Continual Learning}
Classic CL can be divided into several categories including replay-based methods~\cite{ratcliff1990connectionist,derpp,clser,wu2024meta,wang2024dual}, regularization-based methods~\cite{LwF, EWC}, optimization-based methods~\cite{GEM,AGEM,MER,wu2024mitigating}, and LoRA/prompt-based methods~\cite{wu2025sd-lora,zhiqi2025advancing,piao2024federated,yang2024continual,wang2025singular}. Among these, replay-based methods stand out for their competitive performance by using a memory buffer to store a small part of samples from previous tasks.
One important class within replay-based methods~\cite{derpp,clser} focuses on maximizing the utility of stored samples. For example, DER++~\cite{derpp} uses knowledge distillation to effectively utilize old task data stored in the buffer, while CLSER~\cite{clser} integrates a fast learning module to adapt to new tasks and a slow learning module to ensure stable integration of prior knowledge. 

Other critical approaches~\cite{icarl,GSSaljundi2019gradient,OCSyoon2021online} focus on optimizing the quality of data stored in the memory buffer through coreset selection, which prioritizes more representative and higher-quality samples for replay. For instance, iCaRL~\cite{icarl} employs a nearest sample mean rule for classification, and RM~\cite{RMbang2021rainbow} uses interval selection based on classification uncertainty~\cite{uncertaintygal2016dropout}. OCS~\cite{OCSyoon2021online} selects coresets based on mini-batch gradient similarity and inter-batch gradient diversity. However, these methods rely on manually designed strategies for coreset selection, which may limit their performance in complex scenarios. To address this limitation, some approaches aim to automate coreset selection. For instance, a bi-level optimization approach~\cite{borsos2020coresets} dynamically adjusts coreset selection based on model performance. GCR~\cite{GCRtiwari2022gcr} introduces gradient approximation to reduce computational costs, while BCSR~\cite{BCSRhao2024bilevel} improves upon~\cite{zhou2022probabilistic} by incorporating a smoothed top-K regularization term. Compared to these methods, our method considers a learnable memory buffer that distills global information from all seen tasks, thus preserving more important knowledge.

\subsection{Dataset Distillation}
Dataset distillation seeks to condense a large-scale dataset into a smaller synthetic dataset, such that a model trained on this condensed dataset performs comparably to one trained on the full dataset~\cite{DDwang2018dataset}. This process involves optimizing the synthetic dataset to approximate the training dynamics of the full dataset, enabling the distilled data to achieve similar effects with fewer training steps.

Gradient-based methods~\cite{DCzhao2020dataset, zhao2021dataset, lee2022dataset, jiang2023delving, kim2022dataset, zhang2023accelerating} focus on aligning the gradients between the original and synthetic datasets during training. For example, DC~\cite{DCzhao2020dataset} formulates the objective as a gradient matching problem, while DCC~\cite{lee2022dataset} introduces a new gradient matching distance metric that incorporates class-discriminative features. MTT~\cite{MTTcazenavette2022dataset}, a multi-step parameter-based method~\cite{li2024dataset, cui2023scaling}, tracks the training trajectory over multiple steps to improve alignment. On the other hand, distribution-based methods focus on aligning feature statistics between the synthetic and original datasets. For instance, DM~\cite{DMzhao2023dataset} employs Maximum Mean Discrepancy~\cite{mmdgretton2012kernel} to match feature distributions, while \cite{wang2022cafe} ensures consistency in feature statistics extracted by each network layer for synthetic and real samples. Despite their effectiveness, existing dataset distillation methods still face challenges in replay-based CL. Specifically, distillation samples from old tasks may be lost during memory buffer updates, disrupting the data distribution. Moreover, directly optimizing synthetic samples incurs significant computational costs. To address these issues, we propose a continual label distillation process that leverages a bi-level optimization mechanism. This approach preserves data distribution and reduces computational overhead, offering a more efficient solution for CL.

\section{Preliminaries}
Consider a classification model $f_\theta$ trained on $N$ sequentially arriving tasks, denoted as $\{\mathcal{T}^1,\mathcal{T}^2,...,\mathcal{T}^N\}$, where the $n$-th task is represented as $\mathcal{T}^n=\{X^n,Y^n\}$, with $X^n$ and $Y^n$ being the input data and corresponding labels, respectively. For the $n$-th task, the ideal objective of CL is to minimize the following empirical risk,
\begin{equation}
    \theta^*  = \arg\min_{\theta}  \mathcal{L}(f_{\theta};\mathcal T^{[1:n]}),
    \label{eqn:original-obj}
\end{equation}
where $\theta^*$ denote the optimal model weights, $\mathcal{L}$ refer to the cross-entropy loss, and $\mathcal T^{[1:n]}=\cup_{k=1}^n \mathcal{T}^k$ means the all seen $n$ tasks.
When learning the $n$-th task, memory constraints and privacy concerns restrict access to the complete set of examples from the previous tasks $\mathcal{T}^{1:n-1}$. To mitigate this limitation, replay-based CL methods leverage a small memory buffer $\mathcal M^{n-1}=\mathcal E^{[1:n-1]}$, where $\mathcal E^k\subset\mathcal T^k$ (with $|\mathcal E^k|\ll|\mathcal T^k|, k\in[n-1]$) is a subset of examples sampled from $\mathcal{T}^{k}$. Using this buffer, the training objective in Eqn.~(\ref{eqn:original-obj}) can be approximated as follows,
\begin{equation}
    \theta^*  = \arg\min_\theta  \mathcal{L}(f_{\theta};\mathcal M^{n-1} \cup \mathcal{T}^n).
    \label{core-set}
\end{equation}

\textbf{Dataset distillation in CL.} Similar to traditional reply-based methods, recent distillation methods \cite{liu2020mnemonics,LoDM} also maintain a small memory buffer $\mathcal M^{n-1}=\mathcal E^{[1:n-1]}$ to reply when learning task $n$. In contrast, these methods distill each $\mathcal E^k$ from $\mathcal T^k$ at the timestamp $k$. Specifically, this can be formulated as the following bi-level optimization problem:
\begin{equation}
\begin{aligned} 
    \mathcal E^k &=\arg\min_{\mathcal{E}^k_{tmp}} \mathcal{L}(f_{\theta^*(\mathcal E_{tmp}^k)};\mathcal T^k) \\
    \quad  \text{s.t.} & \quad\theta^*(\mathcal E^k_{tmp}) = \arg\min_{\theta} \mathcal{L}(f_{\theta} ;\mathcal E^k_{tmp}),
    \label{eqn:Datadistillation}
\end{aligned}
\end{equation}
where the inner loop trains a temporary classifier model $f_{\theta}$ on the synthetic data $\mathcal E^k_{tmp}$ (treated as a hyper-parameter), and the outer loop uses the entire data $\mathcal T^k$ as a validation set to optimize $\mathcal E^k_{tmp}$. It shows that, each $\mathcal E^k$ in $\mathcal M^{n-1}$ only condenses intra-task information within task $k$, such that the resulting buffer neglects the relationships between different tasks. To illustrate this, we consider a three-task CL scenario and demonstrate the evolution of the memory buffer in Fig. \ref{fig:framework}(a). Throughout the CL process, we can see that the subsets from different tasks do not interact, implying that they are treated as independently distributed across tasks.

\section{Method}
\label{sec:method}
\begin{figure*}[t]
\begin{center}
\centerline{\includegraphics[width=0.75\textwidth]{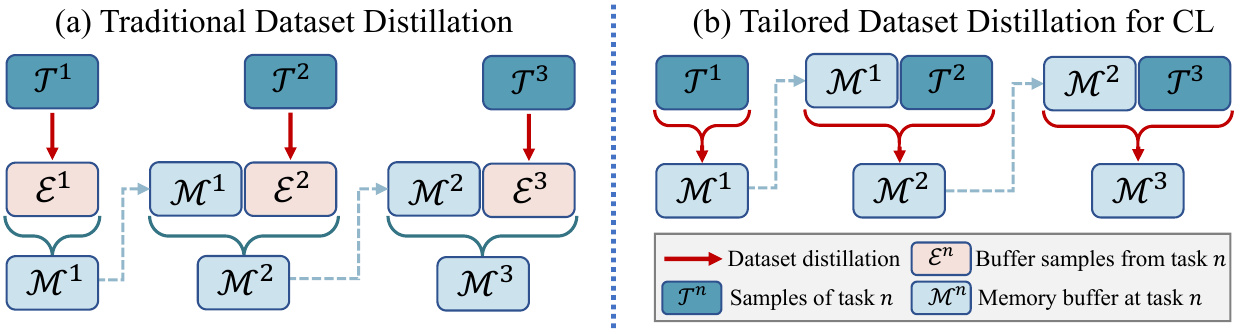}}
\vspace{-2mm}
\caption{Comparison between traditional dataset distillation~(a) and our proposed method~(b). At the $n$-th task $\mathcal{T}^n$, the distilled data $\mathcal{E}^n$ in the traditional approach contains only task-specific information from $\mathcal{T}^n$, while the samples in $\mathcal{M}^n$ fail to capture inter-task relationships as illustrate in ~(a). In contrast, our method distills samples for the memory buffer $\mathcal{M}^n$ from both the current task $\mathcal{T}^n$ and the previous buffer $\mathcal{M}^{n-1}$, thereby preserving correlations across tasks as shown in (b). }
\label{fig:framework}
\end{center}
\vspace{-10mm}
\end{figure*}

In this section, we propose a new dataset distillation framework tailored for CL. Rather than distilling information for individual tasks, as described in Eqn.~(\ref{eqn:Datadistillation}) and Fig. \ref{fig:framework}(a), our approach explicitly captures inter-task relationships.

\vspace{-2mm}
\subsection{Tailored Distillation Framework for CL}
\label{sec: overall_framwork}
Unlike traditional dataset distillation methods exclusively for individual tasks, our proposed framework maintains a learnable memory buffer to distill the global information from
the current task data and accumulated knowledge preserved in the previous memory buffer. As a result, it can effectively capture the relationships among sequential tasks, making it highly suitable for continual learning.

Specifically, when learning the $n$-th task $\mathcal{T}^n$, we directly parameterize the memory buffer $\mathcal M^n_{tmp}$ as learnable parameters, which are then utilized to distill both the current task data $\mathcal T^{n}$ and the existing buffer $\mathcal M^{n-1}$. 
This process is realized through a bi-level optimization strategy, as follows:
\begin{equation}
\begin{aligned} 
\mathcal{M}^{n} &= \arg\min_{\mathcal M^n_{tmp}} \mathcal{L}(f_{\theta^*(\mathcal M^n_{tmp})};\mathcal M^{n-1} \cup\mathcal{T}^n) \\
    \quad  \text{s.t.} & \quad\theta^*(\mathcal M^n_{tmp}) = \arg\min_{\theta} \mathcal{L}(f_{\theta} ; \mathcal M^n_{tmp}),
    \label{eqn:ours bi-level}
\end{aligned}
\end{equation}
where the inner loop trains a temporary classifier $f_{\theta}$ on the learnable buffer $\mathcal M^n_{tmp}$, which is treated as a hyperparameter, to obtain an optimal parameter $\theta^*$, and the outer loop utilizes the combined global information $\mathcal M^{n-1}\cup \mathcal T^n$ from all seen $n$ task as a validation set to optimize $\mathcal M^n_{tmp}$. By optimizing Eqn. (\ref{eqn:ours bi-level}), we obtain the desired buffer $\mathcal M^n$, which corresponds to the optimal $\mathcal M^n_{tmp}$.

As illustrated in Fig.~\ref{fig:framework}~(b), our proposed continuous distillation strategy, described in Eqn.~(\ref{eqn:ours bi-level}), distills $\mathcal{M}^n$ by considering not only the current task and previously learned tasks but also the relationships between them after completing the training of each task. 
This approach avoids the loss of important distilled samples in the buffer $\mathcal{M}$ during updates through reservoir sampling, as illustrated in Fig.\ref{fig:framework}(a). By distilling the buffer and the current task simultaneously, rather than randomly replacing the samples stored in $\mathcal{M}$, we can preserve the core information from both sources while effectively extracting their global task relationships, which are empirically demonstrated to be critical in Sec.~\ref{sec:exp}.

\subsection{Efficient Distillation with Data-Distill-Net}
\label{subsec:ddn}
To capture global information from cross-task data, Eqn. (\ref{eqn:ours bi-level}) parametrizes the entire memory buffer as differentiable parameters during each continual learning phase. However, this strategy introduces significant computational overhead and increases the risk of overfitting during buffer optimization.
To address these challenges, we further propose a lightweight distillation module, Data-Distill-Net (DDN), to ensure effective distillation of global information across tasks while mitigating the computational overhead and overfitting risks associated with full buffer parameterization. For simplicity, we denote the memory buffer as $\mathcal M^{n}=\{X_{buf}^n,Y_{buf}^n\}$ with $Y_{buf}^n$ representing the sets of one-hot labels corresponding to $X_{buf}^n$.

Instead of directly parameterizing the entire $\mathcal{M}^n$ in Eqn.~(\ref{eqn:ours bi-level}), the DDN module first constructs $\mathcal{M}^n$ using existing replay-based methods, such as the commonly used reservoir sampling \cite{Reservoirsamplingvitter1985random}. It then treats the \textit{soft} version of $Y^n_{buf}$ as learnable parameters and employs a unified hyper-network $\mathcal{G}_\omega^n$ to generate these parameters, where $\omega$ denotes its parameters and \(n\) represents the current CL phase. Specifically, this can be formulated as
\begin{equation}
    \tilde{Y}^n_{buf}(\omega) = \mathcal G^n_\omega(X^n_{buf}).
    \label{soft-labels}
\end{equation}
Furthermore, the parameter $\omega$ of $\mathcal{G}_{\omega}^n$ is treated as a hyper-parameter and optimized through a bi-level optimization problem similar to Eqn.~(\ref{eqn:ours bi-level}). It can be expressed as
\begin{equation}
\begin{aligned} 
\omega^*  &= \arg\min_{\omega} \mathcal{L}\bigl(f_{\theta^*(\omega)};\{X_{buf}^{n-1},Y_{buf}^{n-1}\} \cup\mathcal{T}^n\bigr) \\
\quad  \text{s.t.} & \quad\theta^*(\omega) = \arg\min_{\theta} \mathcal{L}\bigl(f_\theta; \{X^n_{buf},\tilde{Y}^n_{buf}(\omega)\}\bigr).
\label{eqn:ddn}
\end{aligned}
\end{equation}
Once the optimal $\omega^*$ is obtained after learning $\mathcal{T}^n$, we only need to retain $X_{buf}^n$ and the lightweight $\mathcal{G}_{\omega^*}^n$ in the buffer, which are subsequently utilized to replay global information from the preceding $n$ tasks, by forming a learnable buffer $\mathcal M^n=\bigl\{{X_{buf}^n, \mathcal G_{\omega^*}^n(X_{buf}^n)}\bigr\}$, during the learning of $\mathcal{T}^{n+1}$.

The success of our proposed distillation framework relies on Data-Distill-Net’s ability to extract global information from \(\mathcal{T}^n \cup \mathcal{M}^{n-1}\). While it employs a hypernetwork-driven parameterization to generate adaptive soft-label distributions for the memory buffer, rather than direct parametric optimization as in Eqn.~(\ref{eqn:ours bi-level}), we have theoretically proven its functional equivalence to the direct parameterization through gradient signal alignment. The detailed proof is provided in Sec.~\ref{sec:theorey}.

\subsection{Data-Distill-Net as a Plugin for Replay-based CL}
\begin{figure}[!t]
\begin{center}
\centerline{\includegraphics[width=0.45\textwidth]{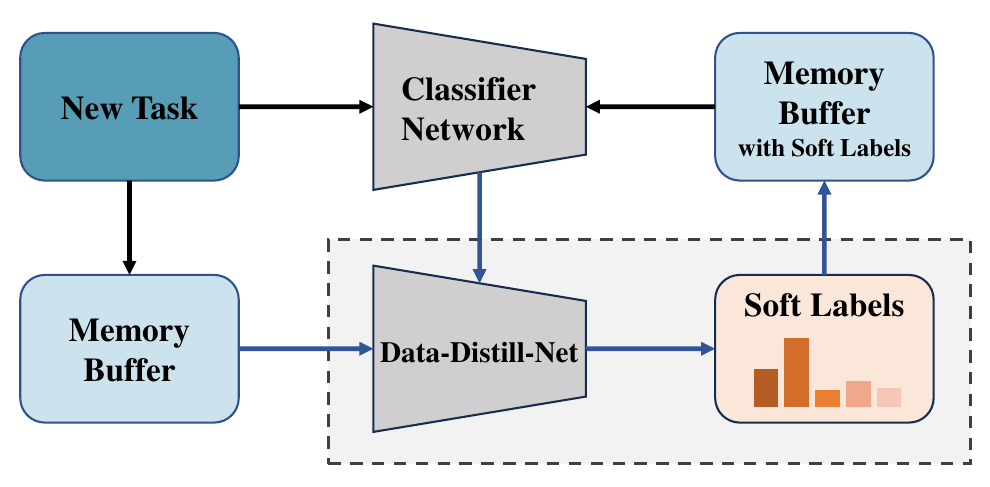}}
\vspace{-3mm}
\caption{Overall framework of the proposed DDN. By generating refined soft labels for buffer samples, DDN enhances the classifier's training, thereby mitigating catastrophic forgetting of prior task knowledge. The training algorithms for the classifier and DDN are detailed in Algorithm~\ref{alg:whole process} and Algorithm~\ref{alg:slg}, respectively.}
\label{fig:pipeline}
\end{center}
\vspace{-6.5mm}
\end{figure}
As aforementioned, our proposed distillation framework does not change the construction or storage manner of the memory buffer data. Instead, it merely introduces a lightweight DDN to condense global information. Therefore, this distillation module can be integrated as a plug-in with any existing replay-based CL method, as illustrated in Fig. \ref{fig:pipeline} and Algorithm \ref{alg:whole process}.

Specifically, during the learning of $\mathcal{T}^n$, given the retained buffer data $X_{buf}^{n-1}$ and the DDN module $\mathcal G_\omega^{n-1}$, we first apply DDN to generate soft labels $\tilde{Y}_{buf}^{n-1}(\omega)=\mathcal G_\omega^{n-1}(X_{buf}^{n-1})$ for these buffer data. We then update the classifier network $f_\theta$ using the resulting learnable buffer $\mathcal M^{n-1}=\{X_{buf}^{n-1}, \tilde{Y}_{buf}^{n-1}\}$ by minimizing the following empirical risk:
\begin{equation}
  \mathcal{L}(f_\theta;\mathcal{T}^n  \cup \mathcal{M}^{n-1}) = \mathcal{L}( f_\theta;\mathcal{T}^n) + \alpha \mathcal{L}( f_\theta;\mathcal{M}^{n-1}),
 \label{hard and soft}
\end{equation}
where $\alpha$ is a hyper-parameter controlling the trade-off between learning new tasks and retaining previously acquired knowledge. Subsequently, we optimize the bi-level optimization problem Eqn. (\ref{eqn:ddn}) to update the DDN module. Due to space constraints, the optimization procedure of Eqn. (\ref{eqn:ddn}) is detailed in Appendix A.

Note that DDN is only enabled during the training phase. During testing, the trained classifier network is used directly for predictions, so there is no additional computational burden during testing.

\textbf{Architectural designs of DDN.} The proposed DDN $\mathcal{G}_\omega^{n}$ is implemented as a Multi-Layer Perceptron (MLP) with two hidden layers, each comprising 200 units followed by a ReLU activation function. For an input $x$, the DDN takes the predicted probabilities from classifier network $f_\theta$ (with gradients stopped) and outputs a soft label $\tilde{y}$. As the inputs of $\mathcal{G}_\omega$ depends on the classifier network, it might make $\mathcal{G}_\omega$ suffer from forgetting when encountering new tasks. To address this issue, we employ an Exponential Moving Average (EMA) to generate soft labels by $\mathcal{G}_\omega$. Specifically, at the end of the $(n-1)$-th task, we save $\mathcal{G}_\omega^{n-1}$ as $\mathcal{G}_{old}$, which is used as a `cached checkpoint' to smooth the outputs of $\mathcal{G}_\omega^{n}$ in the current $n$-th task. Consequently, the soft label generated by DDN in Eqn.~(\ref{soft-labels}) can be reformulated as
\begin{equation}
    \tilde{Y}^{n}_{buf}(\omega) 
    = (1 - \beta) \mathcal{G}^{n}_\omega(X^{n}_{buf})
    + \beta \mathcal{G}_{old}(X^{n}_{buf}),
    \label{EMA soft labels}
\end{equation}
where $\beta \in [0,1]$ is the EMA weight of $\mathcal{G}_{old}$.

In this work, we additionally add the corresponding one-hot label to $\tilde{y}$ and then apply normalization to ensure that it constitutes a valid probability distribution.
Despite its simple structure, DDN leverages its universal approximation capability \cite{mlphaykin1998neural} to effectively capture and represent complex relationships between different classes.

\begin{algorithm}[t]
   \caption{Training of classifier network}
   \label{alg:whole process}
\begin{algorithmic}
   \STATE {\bfseries Input:} current task $\mathcal{T}^n$, classifier network $f_{\theta}$, DDN $\mathcal{G}_\omega^{n-1}$, memory buffer $\mathcal{M}^{n-1}$, max iterations K
   \FOR{$i = 0$ to K} 
   \STATE  $\mathcal{B}^{n}\!=\!\{X^n, Y^n\} \!\leftarrow\!$ SampleMiniBatch($\mathcal{T}^{n}$)
   \STATE $\mathcal{B}^{n-1}_{buf}\!=\!\{X^{n-1}_{buf}, Y^{n-1}_{buf}\} \!\leftarrow\!$ SampleMiniBatch($\mathcal{M}^{n-1}$)
   \STATE Generate soft-labels $\tilde{Y}_{buf}^{n-1}=\mathcal G_\omega^{n-1}(X_{buf}^{n-1})$
   \STATE Update $\theta$ of classifier network by minimize Eq. (\ref{hard and soft})
    \STATE Update DDN by Algorithm $\ref{alg:slg}$
   \ENDFOR
\end{algorithmic}
\end{algorithm}

\begin{algorithm}[t]
   \caption{Training of DDN}
   \label{alg:slg}
\begin{algorithmic}
   \STATE {\bfseries Input:} current task $\mathcal{T}^n$, memory buffer $\mathcal{M}^{n-1}$, max iterations K
   \FOR{$i=0$ to K}
   \STATE \textit{\textcolor{gray}{\# Inner loop}}
  
   \STATE $\mathcal{B}^{in}\!=\!\{X^{in},Y^{in}\} \!\leftarrow\!$ SampleMiniBatch($\mathcal{M}^n$)
   \STATE Generate soft labels $\tilde{Y}^{in}(\omega)=\mathcal G_\omega^{n-1}(X^{in})$
   \STATE Update $\theta(\omega)$ of classifier network by minimizing  $\mathcal{L}(f_\theta;\{X^{in},Y^{in}\}) + \alpha \cdot \mathcal{L}(f_\theta;\{X^{in},\tilde{Y}^{in}(\omega)\})$ 
   \STATE \textit{\textcolor{gray}{\# Outer loop}}
   \STATE $\mathcal{B}^{out}\!=\!\{X^{out},Y^{out
   }\} \!\!\leftarrow\!\!$ SampleMiniBatch($\mathcal{M}^{n-1} \!\cup\! \mathcal{T}^n$)
   \STATE  Update $\omega$ of DDN by minimizing $\mathcal{L}(f_{\theta(\omega)};\mathcal{B}^{out})$ 
   \ENDFOR
\end{algorithmic}
\end{algorithm}

\section{Theoretical Analysis}
\label{sec:theorey}
In this section, we establish the theoretical connections between two memory buffer parameterization paradigms. Whereas Eqn.~(\ref{eqn:ours bi-level}) implements direct parametric optimization by treating the entire memory buffer as learnable parameters, Eqn.~(\ref{eqn:ddn}) proposes a hypernetwork-driven parameterization that generates adaptive soft-label distributions for memory entries. 
To formalize this equivalence, we first demonstrate that our bi-level framework Eqn.~(\ref{eqn:ours bi-level}) corresponds to the gradient-matching formulation below.  

\begin{theorem}
Let $\nabla_\theta \mathcal{L}(f_\theta; \mathcal{M}^n)$  be the gradients w.r.t. $\theta$ on the parameterized memory buffer $\mathcal M^n$, and $\nabla_\theta \mathcal{L}(f_\theta; \mathcal{M}^{n-1} \cup \mathcal{T}^n)$ be the gradients w.r.t. $\theta$ on the current task $\mathcal T^n$ and the previous buffer $\mathcal M^{n-1}$. The bi-level framework in Eqn.~(\ref{eqn:ours bi-level}) is equivalent to minimizing the following empirical risk:
\begin{equation}
\min_{\mathcal M^n}-\bigl\langle \nabla_\theta \mathcal{L}(\theta; \mathcal M^n),\nabla_\theta \mathcal{L}(\theta; \mathcal{M}^{n-1} \cup \mathcal{T}^n) \bigr\rangle.
\label{eqn:theorem_1}
\end{equation}
\label{thm:bi_level_ours}
\vspace{-4mm}
\end{theorem}
Next, we show that the bi-level framework in Eqn.~(\ref{eqn:ddn}) is equivalent to the following gradient-matching formulation.
\begin{theorem}
Let $\nabla_\theta \mathcal{L}(f_\theta; \mathcal{M}^n)$ be the gradients w.r.t. $\theta$ on the learnable memory buffer $\mathcal M^n=\bigl\{{X_{buf}^n, \mathcal G_{\omega}^n(X_{buf}^n)}\bigr\}$, and $\nabla_\theta \mathcal{L}(f_\theta; \mathcal{M}^{n-1} \cup \mathcal{T}^n)$ be the gradients w.r.t. $\theta$ on the current task $\mathcal T^n$ and the previous buffer $\mathcal M^{n-1}$. The bi-level framework in Eqn.~(\ref{eqn:ddn}) is equivalent to minimizing the following empirical risk:
\begin{equation}
\min_{\omega}-\bigl\langle \nabla_\theta \mathcal{L}(\theta; \mathcal M^n),\nabla_\theta \mathcal{L}(\theta; \mathcal{M}^{n-1} \cup \mathcal{T}^n) \bigr\rangle.
\label{eqn:theorem_2}
\end{equation}
\label{thm:bi_level_ours2}
\vspace{-6mm}
\end{theorem}
Theorem~\ref{thm:bi_level_ours} and ~\ref{thm:bi_level_ours2} demonstrate that the direct parametric optimization approach in Eqn.~(\ref{eqn:ours bi-level}) and the hypernetwork-driven parameterization method in Eqn.~(\ref{eqn:ddn}) play analogous roles throughout the distillation process. Specifically, once both optimization objectives achieve their optimums, we have $\nabla_\theta \mathcal{L}\bigl(f_\theta; \{X_{buf}^n, \mathcal{G}^n_\omega\}\bigr) \simeq \nabla_\theta \mathcal{L}(\theta; \mathcal{M}^{n-1} \cup \mathcal{T}^n) \simeq \nabla_\theta \mathcal{L}\bigl(f_\theta; \mathcal{M}^n\bigr)$. This demonstrates that our method significantly reduces computational overhead by leveraging hypernetwork-driven parameterization instead of direct parametric optimization while still achieving the same optimum. 

\begin{table*}[t]
\caption{Comparison results on three datasets under the online continual learning setting, evaluated by ACC and FM. \colorbox[HTML]{D8D8D8}{Gray cells} represent our method applied to four baselines.}
\vspace{-3mm}
\label{tab:online_comparison}
\begin{center}
\centering
\resizebox{\textwidth}{!}{
\begin{tabular}{lcccccccccccc}
\hline
\multicolumn{1}{c}{} & \multicolumn{4}{c}{Split CIFAR-10} & \multicolumn{4}{c}{Split CIFAR-100} & \multicolumn{4}{c}{Split Tiny-ImageNet} \\ \cline{2-13} 
\multicolumn{1}{c}{} & \multicolumn{2}{c}{M=0.2K} & \multicolumn{2}{c}{M=0.5K} & \multicolumn{2}{c}{M=2K} & \multicolumn{2}{c}{M=5K} & \multicolumn{2}{c}{M=2K} & \multicolumn{2}{c}{M=5K} \\ \cline{2-13} 
\multicolumn{1}{c}{\multirow{-3}{*}{Method}} & ACC $\uparrow$ & FM $\downarrow$ & ACC $\uparrow$ & FM $\downarrow$ & ACC $\uparrow$ & FM $\downarrow$ & ACC $\uparrow$ & FM $\downarrow$ & ACC $\uparrow$ & FM $\downarrow$ & ACC $\uparrow$ & FM $\downarrow$ \\ \hline
\multicolumn{1}{c}{LUCIR~\cite{LUCIR}} & 23.59 & 35.59 & 24.63 & 31.89 & 8.28 & 16.07 & 12.31 & 14.02 & 4.47 & 20.40 & 5.29 & 20.28 \\
\multicolumn{1}{c}{iCaRL~\cite{icarl}} & 40.49 & 26.84 & 44.50 & 24.87 & 9.13 & 7.79 & 9.13 & 8.14 & 4.03 & 4.93 & 4.03 & 5.15 \\
\multicolumn{1}{c}{BiC~\cite{BIC}} & 27.71 & 66.45 & 35.47 & 47.92 & 16.32 & 36.70 & 20.89 & 32.33 & 5.43 & 40.14 & 7.50 & 38.52 \\ \hline
\multicolumn{1}{c}{Mnemonics~\cite{liu2020mnemonics}} & 37.08 & 38.27 & 43.12 & 29.88 & 22.93 & 27.29 & 23.82 & 28.05 & 15.38 & 26.72 & 16.91 & 24.98 \\
\multicolumn{1}{c}{LoDM~\cite{LoDM}} & 30.57 & 17.84 & 29.22 & 19.57 & 14.78 & 11.20 & 15.02 & 10.95 & 4.43 & 9.08 & 4.83 & 10.72 \\ \hline
\multicolumn{1}{c}{ER~\cite{ERratcliff1990connectionist}} & 36.05 & 49.99 & 44.34 & 36.90 & 21.23 & 33.99 & 23.78 & 33.1 & 15.35 & 31.45 & 16.86 & 30.76 \\
\rowcolor[HTML]{D8D8D8}
\multicolumn{1}{c}{ER DDN~(ours)} & 41.93 & 35.28 & 48.11 & 30.07 & 23.41 & 30.08 & 24.48 & 30.25 & 15.13 & 28.06 & 17.14 & 28.9 \\ \hline
\multicolumn{1}{c}{DER++~\cite{derpp}} & 41.54 & 39.49 & 49.11 & 34.24 & 17.90 & 43.76 & 17.95 & 44.54 & 11.64 & 40.69 & 11.47 & 41.44 \\
\rowcolor[HTML]{D8D8D8}
\multicolumn{1}{c}{DER++ DDN~(ours)} & 42.82 & 37.26 & 49.48 & 28.18 & 21.92 & 35.59 & 22.29 & 37.22 & 13.84 & 35.46 & 15.54 & 34.44 \\ \hline
\multicolumn{1}{c}{CLSER~\cite{clser}} & 39.03 & 48.49 & 44.58 & 37.31 & 21.65 & 36.15 & 23.40 & 35.38 & 14.79 & 33.19 & 16.64 & 32.79 \\
\rowcolor[HTML]{D8D8D8}
\multicolumn{1}{c}{CLSER DDN~(ours)} & 40.95 & 40.06 & 47.37 & 29.74 & 22.80 & 31.48 & 25.05 & 30.44 & 15.16 & 32.87 & 17.61 & 28.93 \\ \hline
\multicolumn{1}{c}{ER-ACE~\cite{ERACE}} & 42.81 & 17.32 & 47.02 & 17.53 & 25.74 & 7.26 & 28.53 & 6.13 & 17.64 & 7.58 & 20.37 & 5.53 \\
\rowcolor[HTML]{D8D8D8}
\multicolumn{1}{c}{ER-ACE DDN~(ours)} & 43.37 & 17.60 & 50.02 & 14.50 & 25.94 & 8.71 & 28.73 & 7.73 & 18.25 & 9.81 & 21.02 & 7.02\\\hline
\end{tabular}
}
\end{center}
\vspace{-5mm}
\end{table*}

\vspace{-2mm}
\section{Experiments}
\label{sec:exp}
To verify the effectiveness of our method DDN, we compared it with several current state-of-the-art CL baselines under different experimental setups including online and offline settings. Additionally, we conducted a series of ablation experiments to demonstrate the rationale of our method.
\vspace{-6mm}
\subsection{Experimental Settings}
\label{subsec:exp_setting}
\textbf{Datasets}: Our experiments are conducted on three commonly used CL datasets following~\cite{derpp}: Split CIFAR-10~\cite{cifar10}, Split CIFAR-100, and Split Tiny-ImageNet~\cite{tinyimg}. Specifically, we split CIFAR-10 into 5 binary classification tasks. Both CIFAR-100 and Tiny-ImageNet are divided into 10 tasks, with 10 classes and 20 classes per task, respectively. See Appendix C for details.

\textbf{Metrics}:
To fairly evaluate our approach and other comparison methods, we employ the following two metrics in all experiments.
\begin{itemize}[leftmargin=4mm, itemsep=0mm, topsep=-0.5 mm]
    \item \textbf{Average Accuracy} (ACC $\uparrow$):
    The average accuracy measures the overall CL classification ability, \textit{i.e.}, $\mathrm{ACC} = \frac{1}{T} \sum_{i=t}^T A_{t, T}$, where $A_{t, T}$ represents the test accuracy of the model on the $t$-th task $\mathcal{T}^t$ after the training of all $T$ tasks. Additionally, $\uparrow$ indicates that higher is better.
    \item \textbf{Forgetting Measure} (FM $\downarrow$):
    FM measures the average forgetting of each task, where the lower FM ($\downarrow$) is better. Specifically, FM calculates the average difference between the best performance during the whole CL stage $A^*_t$ and the final performance $A_{t, T}$, \textit{i.e.}, $\mathrm{FM} = \frac{1}{T} \sum_{i=t}^T A^*_t - A_{t, T}$.
\end{itemize}

\textbf{Comparison methods}:
The proposed DDN can be easily plugged into various rehearsal-based CL methods and further improve their performance, such as ER~\cite{ERratcliff1990connectionist}, DER++~\cite{derpp}, CLSER~\cite{clser}, and ER-ACE~\cite{ERACE}. To verify the strength of our method, we also compare the performance with several representative rehearsal methods, \textit{i.e.}, LUCIR~\cite{LUCIR}, iCaRL~\cite{icarl}, BiC~\cite{BIC}, and two dataset distillation CL methods LoDM~\cite{LoDM} and Mnemonics~\cite{liu2020mnemonics}, whose memory buffer applied distillation samples. Details of comparison methods please refer to Appendix C.

\textbf{Implementation details}:
We focus primarily on online and offline CL settings. Specifically, in \textbf{online CL}, data arrives sequentially, and each sample is trained only once, except for those stored in a memory buffer, making it more challenging as models must learn quickly while mitigating forgetting. In contrast, \textbf{offline CL} allows storing the current task samples and training multiple times, leading to more stable performance, though catastrophic forgetting remains a key challenge.
In our experiments, we take the average results of 10 runs and 3 runs to alleviate random factors for online and offline CL, respectively. We adopt ResNet-18 as backbone follow \cite{derpp} and train all methods by stochastic gradient descent (SGD) optimizer. For our DDN, we use Adam~\cite{adam} optimizer with the learning rate as 0.001 for Split CIFAR-10, 0.01 for Split CIFAR-100, and 0.0001 for Split Tiny-ImageNet. More implementation details are shown in Appendix C.

\vspace{-2mm}
\subsection{Results of Online and Offline CL}
\textbf{Online CL.} Table~\ref{tab:online_comparison} shows the ACC and FM results for our method and all comparison methods. Note that our approach has been applied to four different rehearsal-based CL baselines, marked as `DDN' with gray cells in the table. From Table~\ref{tab:online_comparison}, it is evident that our method improves the classification accuracy of the corresponding baseline while significantly reducing forgetting. For example, our ER DDN improves the ACC of the baseline ER by $5.88\%$ on Split CIFAR-10 with 0.2K memory buffer samples. Even for ER-ACE with relatively lower FM, our method can further improve the classification accuracy from a global perspective and produce a comparable FM. These results demonstrate that the proposed method effectively distills knowledge from all tasks into DDN by generating soft labels, leading to better CL performance.
For a fair comparison with other dataset distillation CL methods, we also applied LoDM and Mnemonics to ER. LoDM performs poorly in online CL due to its reliance on excessive iterations to optimize condensed data samples. Mnemonics, which initializes synthetic data as real images from the training set, offers a slight improvement over ER but still falls short of our method. These results demonstrate that our distillation framework is better suited for CL tasks, significantly reducing computational costs while enhancing baseline performance.

\begin{table*}[!htp]
\caption{Comparison results on three datasets under the offline continual learning setting, evaluated by ACC and FM. \colorbox[HTML]{D8D8D8}{Gray cells} represent our method applied to four baselines.}
\label{tab:offline_comparison}
\vspace{-3mm}
\begin{center}
\resizebox{\textwidth}{!}{ 
\begin{tabular}{lcccccccccccc}
\hline
\multicolumn{1}{c}{} & \multicolumn{4}{c}{Split CIFAR-10} & \multicolumn{4}{c}{Split CIFAR-100} & \multicolumn{4}{c}{Split Tiny-ImageNet} \\ \cline{2-13} 
\multicolumn{1}{c}{} & \multicolumn{2}{c}{M=0.2K} & \multicolumn{2}{c}{M=0.5K} & \multicolumn{2}{c}{M=2K} & \multicolumn{2}{c}{M=5K} & \multicolumn{2}{c}{M=2K} & \multicolumn{2}{c}{M=5K} \\ \cline{2-13} 
\multicolumn{1}{c}{\multirow{-3}{*}{Method}} & ACC $\uparrow$ & FM $\downarrow$ & ACC $\uparrow$ & FM $\downarrow$ & ACC $\uparrow$ & FM $\downarrow$ & ACC $\uparrow$ & FM $\downarrow$ & ACC $\uparrow$ & FM $\downarrow$ & ACC $\uparrow$ & FM $\downarrow$ \\ \hline
\multicolumn{1}{c}{ER~\cite{ERratcliff1990connectionist}} & 54.91 & 48.43 & 65.31 & 35.45 & 39.37 & 53.62 & 50.05 & 40.17 & 18.71 & 64.77 & 27.58 & 54.01 \\
\rowcolor[HTML]{D8D8D8}
\multicolumn{1}{c}{ER DDN~(ours)} & 58.93 & 43.24 & 68.56 & 29.57 & 46.32 & 43.45 & 56.57 & 30.63 & 22.93 & 47.71 & 31.02 & 42.40 \\ \hline
\multicolumn{1}{c}{DER++~\cite{derpp}} & 64.59 & 34.99 & 73.31 & 23.45 & 53.04 & 32.56 & 60.45 & 23.39 & 32.22 & 42.34 & 39.88 & 31.50 \\
\rowcolor[HTML]{D8D8D8}
\multicolumn{1}{c}{DER++ DDN~(ours)} & 65.73 & 31.34 & 73.96 & 20.01 & 52.65 & 33.57 & 59.85 & 25.20 & 32.42 & 40.18 & 39.37 & 30.04 \\ \hline
\multicolumn{1}{c}{CLSER~\cite{clser}} & 55.25 & 47.14 & 69.05 & 28.42 & 44.84 & 46.13 & 54.71 & 34.26 & 23.75 & 58.75 & 30.93 & 50.70 \\
\rowcolor[HTML]{D8D8D8}
\multicolumn{1}{c}{CLSER DDN~(ours)} & 57.34 & 42.00 & 69.79 & 27.88 & 48.46 & 37.34 & 55.85 & 27.76 & 24.97 & 53.69 & 35.58 & 40.95 \\ \hline
\multicolumn{1}{c}{ER-ACE~\cite{ERACE}} & 64.35 & 18.38 & 71.95 & 14.44 & 50.29 & 28.01 & 56.34 & 21.56 & 32.37 & 36.42 & 38.92 & 28.65\\
\rowcolor[HTML]{D8D8D8}
\multicolumn{1}{c}{ER-ACE DDN~(ours)} & 66.12 & 19.25 & 72.20 & 14.96 & 51.44	& 30.91 & 59.76 & 22.44 & 32.19	& 43.23 & 37.67 & 39.37 \\ \hline
\end{tabular}
}
\end{center}
\vspace{-3mm}
\end{table*}

\textbf{Offline CL.} Different from online CL, models in offline CL typically achieve better and more stable results, though the forgetting issue remains pronounced. Table~\ref{tab:offline_comparison} presents the performance of our approach and comparison methods in the offline CL setting. From these results, our method significantly improves ACC and reduces FM across almost all datasets and baselines. For example, on the CIFAR-100 dataset with 2K buffer samples, our method improves by $6.95\%$ compared to the baseline. For ER-ACE, our method also enhances the accuracy with comparable FM. These superior results suggest that our DDN effectively captures and transfers knowledge from all tasks. Therefore, with our distillation framework and buffered soft labels, most rehearsal-based CL methods can be further improved in both online and offline settings.

\subsection{Discussion}
\textbf{Comparison with other soft label methods}:
As aforementioned, we distill information from all tasks into DDN by generating learnable soft lables. To demonstrate the superiority of the distilled soft label, we compare our approach with other soft label methods based on ER in Table~\ref{tab:soft_label}.
First, we create random soft labels by adding uniform noise to hard labels with normalization, referred to as `RANDOM' in Table~\ref{tab:soft_label}. Such random soft labels do not provide additional useful information and even harm baseline performance. Then we evaluate label smoothing~\cite{labelsmoothing} with different smoothing strength, denoted as `ER LS'. This approach suffers from instability during CL training and struggles to improve baseline. In contrast, our method leads to significant improvements, highlighting the effectiveness of the soft labels generated by our distillation framework.

\begin{table}[t]
\centering
\caption{The online CL comparison results of our method and other soft label methods on Split CIFAR-10 with various buffer sizes.}
\label{tab:soft_label}
\vspace{-1mm}
\begin{center}
\resizebox{0.8\columnwidth}{!}{\begin{tabular}{ccccc}
\hline
 & \multicolumn{4}{c}{Split CIFAR-10} \\ \cline{2-5}
 & \multicolumn{2}{c}{M=0.2K} & \multicolumn{2}{c}{M=0.5K} \\ \cline{2-5}
\multirow{-3}{*}{Method} & ACC~$\uparrow$ & FM~$\downarrow$ & ACC~$\uparrow$ & FM~$\downarrow$ \\ \hline
\multicolumn{1}{c}{ER} & 36.05 & 49.99 & 44.34 & 36.90 \\
\multicolumn{1}{c}{RANDOM} & 33.47 & 55.64 & 43.85 & 40.56 \\
\multicolumn{1}{c}{ER LS} ($\epsilon$=0.1) & 35.46 & 50.72 & 45.04 & 37.25 \\
\multicolumn{1}{c}{ER LS} ($\epsilon$=0.5) & 34.24 & 52.96 & 42.59 & 41.87 \\
\multicolumn{1}{c}{ER L2Y} & 37.60 & 42.83 & 44.44 & 37.10 \\
\rowcolor[HTML]{D8D8D8} 
\multicolumn{1}{c}{ER DDN~(ours)} & 41.93 & 35.28 & 48.11 & 30.07 \\ \hline
\end{tabular}}
\end{center}
\vspace{-8mm}
\end{table}

\begin{figure*}[t]
\begin{center}
\centerline{\includegraphics[width=0.95\linewidth]{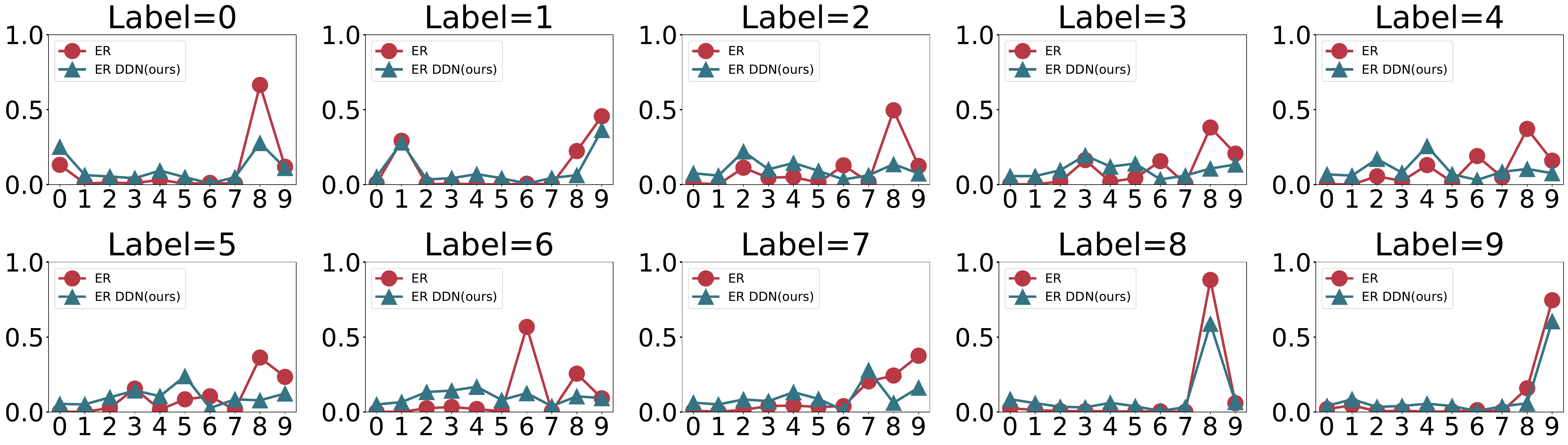}}
\vspace{-3mm}
\caption{Under online CL, comparison of the predicted probability distributions for different class samples between ER and ER DDN(ours) on Split CIFAR-10 with buffer size M = 0.2K.}
\label{fig:prob_distribution}
\end{center}
\vspace{-10mm}
\end{figure*}

Furthermore, to verify the strength of $\mathcal G_\omega$, we directly optimize soft labels of memory buffer samples by our distillation framework as `ER L2Y'. While this method also improves the performance of ER, it cannot outperform our method with the $\mathcal G_\omega$, indicating the importance of this module.

\textbf{Probability distribution predicted by DDN}:
To figure out the effect of our proposed DDN on classifier network training, under the online CL setting, we calculate the probability distribution of each class predicted by ER and our ER DDN after the whole CL training of Split CIFAR-10 with M = 0.2K in Fig.~\ref{fig:prob_distribution}. For the old classes of previous tasks (\textit{i.e.}, Label=$0,\cdots, 7$), the predicted probabilities are seriously biased toward the new classes (\textit{i.e.}, Label=$8,9$). In contrast, our method significantly corrects this bias and enhances the predicted probabilities of the ground truth class. On the other hand, for the new classes (\textit{i.e.}, Label=$8,9$), our proposed ER DDN decreases prediction confidence for these new class samples without loss of classification accuracy.
Therefore, our method can automatically adapt to CL distributions by distilling knowledge from all tasks, and thus accelerate network training to achieve better performance.

\textbf{Effect of small buffer size}:
To further investigate the effectiveness of our method in scenarios with small buffer sizes, we apply the proposed DDN to four baselines on Split CIFAR-10, as shown in Table~\ref{tab:extreme scenarios}. When the buffer size is M = 0.05K, our method ER DDN improves the ACC of the baseline ER by approximately $4.1\%$ and reduces FM by around $12.03\%$. These results demonstrate that our method can effectively extract and distill information from training samples across all tasks into the limited memory buffer, significantly enhancing the performance of the corresponding rehearsal baselines in memory-constrained scenarios.

\begin{table}[t]
\caption{Under online CL, comparison results of our method and corresponding baselines on Split CIFAR-10 with small buffer sizes.}
\label{tab:extreme scenarios}
\vspace{-3mm}
\begin{center}
\resizebox{1.0\linewidth}{!}{
\begin{tabular}{ccccr}
\hline
 & \multicolumn{4}{c}{Split CIFAR-10} \\ \cline{2-5}
 & \multicolumn{2}{c}{M=0.05K} & \multicolumn{2}{c}{M=0.1K} \\ \cline{2-5}
\multirow{-3}{*}{Method} & ACC~$\uparrow$ & FM~$\downarrow$ & ACC~$\uparrow$ & FM~$\downarrow$ \\\hline
\multicolumn{1}{c}{ER~\cite{ERratcliff1990connectionist}} & 22.44 & 74.46 & 28.08 & 61.30 \\
\rowcolor[HTML]{D8D8D8} 
\multicolumn{1}{c}{ER DDN~(ours)} &26.51 & 62.43 & 34.54 & 45.84 \\\hline
\multicolumn{1}{c}{DER++~\cite{derpp}} & 30.11 & 57.91 & 36.21 & 49.66 \\
\rowcolor[HTML]{D8D8D8} 
\multicolumn{1}{c}{DER++ DDN~(ours)} & 31.52 & 51.82 & 37.19 & 45.85 \\\hline
\multicolumn{1}{c}{CLSER~\cite{clser}} & 25.53 & 67.46 & 30.58 & 54.59 \\
\rowcolor[HTML]{D8D8D8} 
\multicolumn{1}{c}{CLSER DDN~(ours)} & 28.08 & 55.93 & 34.77 & 45.90\\\hline
\multicolumn{1}{c}{ER-ACE~\cite{ERACE}} & 33.77 & 22.73 & 36.72 & 21.19 \\
\rowcolor[HTML]{D8D8D8} 
\multicolumn{1}{c}{ER-ACE DDN~(ours)} &34.15 & 21.47 & 39.24 & 19.49 \\\hline
\end{tabular}
}
\end{center}
\vspace{-3mm}
\end{table}

\begin{figure}
    \centering
    \subfigure[]{
        \includegraphics[width=0.46\linewidth]{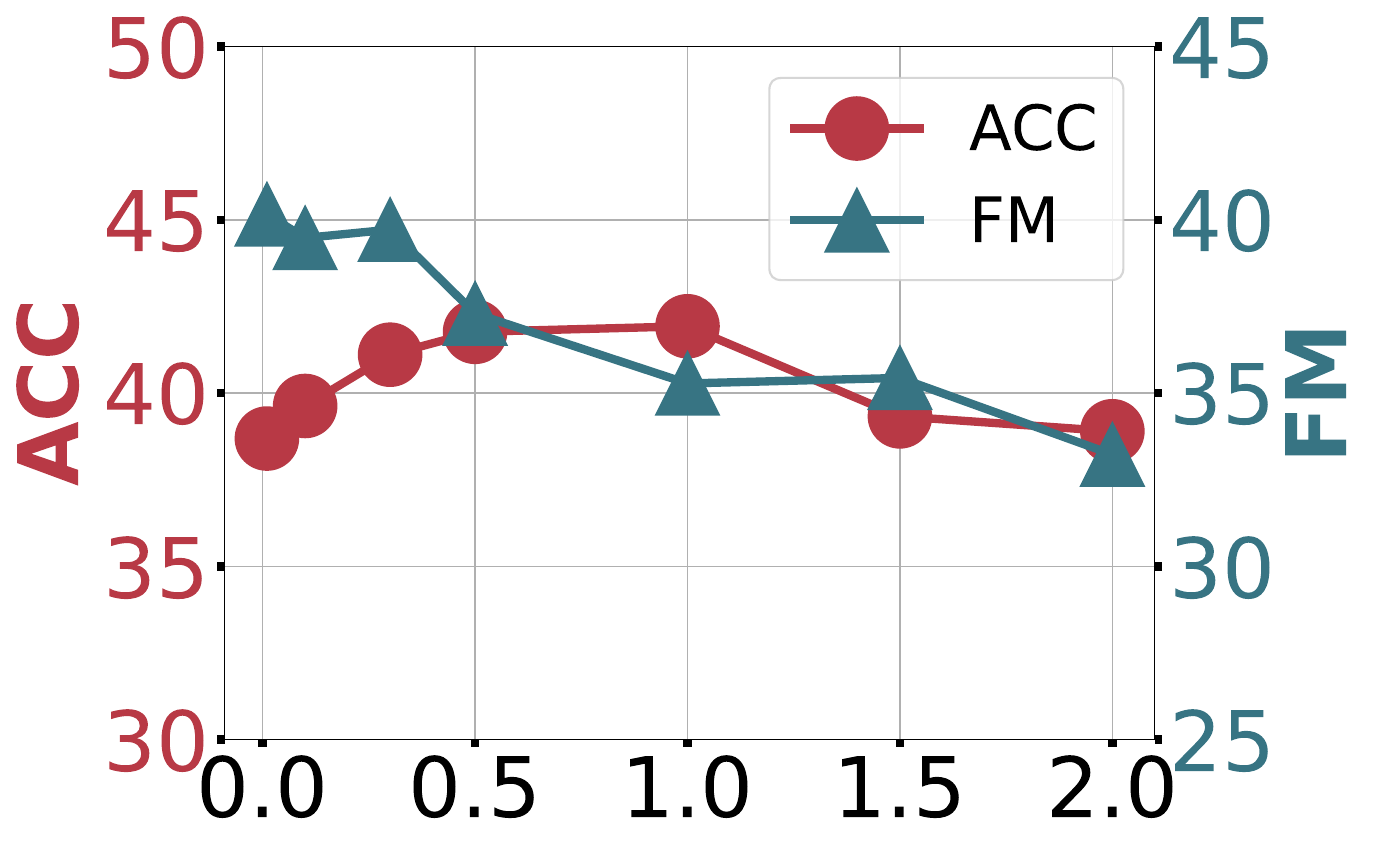}
        \label{subfig:alpha}
    }
    \subfigure[]{
        \includegraphics[width=0.46\linewidth]{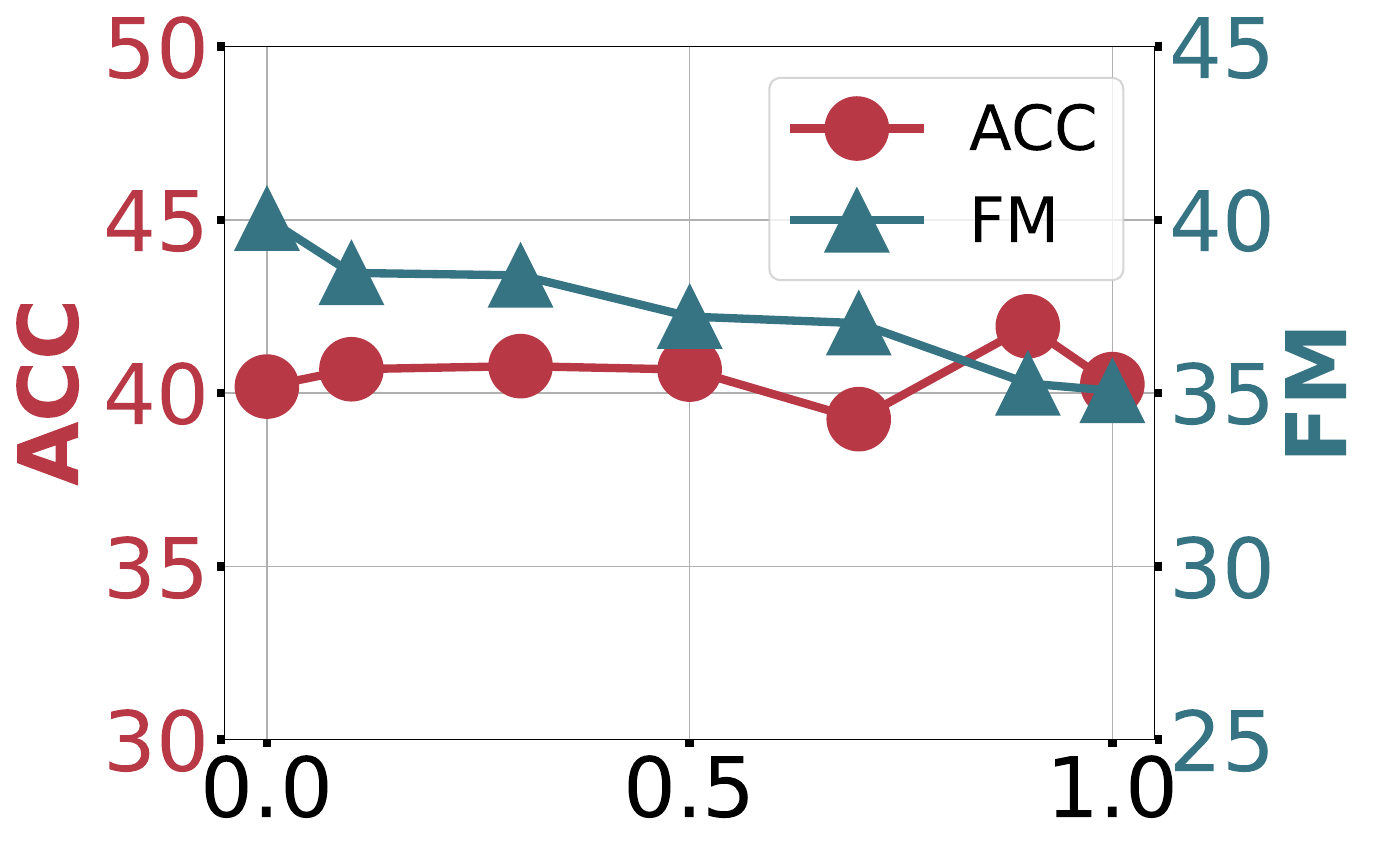}
        \label{subfig:beta}
    }
    \vspace{-6mm}
    \caption{Under online CL,
    ACC and FM of different $\alpha$ and $\beta$ of our method on Split CIFAR-10 with buffer size M = 0.2K. (a) Effect of $\alpha$ with fixed $\beta=0.9$. (b) Effect of $\beta$ with fixed $\alpha=1.0$.}
    \label{fig:alpha_beta}
\vspace{-6mm}
\end{figure}

\textbf{Effect of $\alpha$ and $\beta$}:
In our distillation framework, $\alpha$ and $\beta$ are key hyperparameters to control the strength of DDN and the effect of the old generator in EMA, respectively. In Fig.~\ref{fig:alpha_beta}, we illustrate the effect of varying $\alpha$ and $\beta$ on Split CIFAR-10 with M = 0.2K. Our method achieves the highest ACC and a relatively lower FM at $\alpha=1.0$, indicating that our soft label loss is crucial in the classifier network training. Additionally, a larger $\alpha$ results in lower accuracy, likely because the model still relies on hard labels to provide sufficient confidence for classification.
Furthermore, with $\alpha=1.0$, our method performs best when $\beta=0.9$, suggesting that the old $\mathcal{G}_{old}$ plays an important role in alleviating forgetting and improving performance.

\section{Conclusion}

The previous dataset distillation methods applied in the continual learning (CL) problem typically ignored the correlation between each task. To tackle this challenge, we propose a new dataset distillation framework tailored for CL, which maintains a learnable memory buffer to distill global information from all seen tasks. Furthermore, we parameterize a Data-Distill-Net (DDN) to distill global information by generating learnable soft labels for memory buffer samples, reducing the computational burden and overfitting risks during the CL training and distillation stage. We plug the designed DDN in four rehearsal-based CL baselines and the experiments have demonstrated the effectiveness of DDN under both online and offline settings.

\section*{Impact Statement}
This paper aims to advance the field of Machine Learning. Specifically, we propose a novel distillation method tailored for CL, which offers significant advantages by reducing memory and computational requirements, thereby enhancing the accessibility of machine learning models in resource-constrained environments. On the flip side, these benefits might come with potential ethical and societal challenges if not properly managed. The data compression process might unintentionally exclude rare or minority examples, thereby reinforcing existing biases and diminishing fairness in applications. As a result, it is essential for practitioners to ensure transparency in how distilled datasets are created and evaluated, as well as to conduct rigorous assessments to guarantee that diverse populations are adequately represented.

\bibliography{example_paper}
\bibliographystyle{icml2025}

\newpage
\appendix
\twocolumn[
\begin{center}
    {\LARGE \textbf{Supplementary Materials\\
    Data-Distill-Net: A Data Distillation Approach \\
           Tailored for Reply-based Continual Learning}}
\end{center}
\vskip 0.3in
]
\section{Optimization of Eqn. (\ref{eqn:ddn})}
We first recall Eqn.~(\ref{eqn:ddn}) in the main text, representing our proposed distillation framework tailored for continual learning, which can be represented as follows:
\begin{equation}
\begin{aligned} 
\omega^*  &= \arg\min_{\omega} \mathcal{L}\bigl(f_{\theta^*(\omega)};\{X_{buf}^{n-1},Y_{buf}^{n-1}\} \cup\mathcal{T}^n\bigr) \\
\quad  \text{s.t.} & \quad\theta^*(\omega) = \arg\min_{\theta} \mathcal{L}\bigl(f_\theta; \{X^n_{buf},\tilde{Y}^n_{buf}(\omega)\bigr\}).
\end{aligned}
\nonumber
\end{equation}
where $\tilde{Y}^n_{buf}(\omega)=\mathcal{G}^n_\omega(X^n_{buf})$ and $\mathcal{G}^n_\omega$ indicates our proposed DDN to distill knowledge from all seen tasks. Note that
Eqn. (\ref{eqn:ddn}) form a nested bi-level optimization problem,
where the optimal solutions of $\theta^*$ and $\omega^*$ are interdependent in the inner and outer optimization problems, and it is difficult to obtain a closed-form solution for $\theta$ and $\omega$. To optimize this bi-level optimization framework in neural networks, we propose to use a gradient-based optimization method to approximately update parameters $\theta$ and $\omega$~\cite{maml, wang2023cba}.

Specifically, we iteratively update $\theta$ and $\omega$ as follows:

(1) \textbf{Update $\theta$}: At the $k_{th}$ iteration, the parameters of DDN $\mathcal{G}$ are denoted by $\omega_k$. We fix $\omega_k$ and use a single step of stochastic gradient descent (SGD) to update the parameter $\theta$ of the classifier network $f$. 
Therefore, the updated parameter $\theta$ are a function of $\omega$, \textit{i.e.},
\begin{equation}
    \theta^{k+1}(\omega) = \theta^k - \eta \cdot \partial_{\theta}\mathcal{L}\bigl(f_\theta; \{X^n_{buf},\tilde{Y}^n_{buf}(\omega)\bigr\})
    \label{theta}
\end{equation}
where $\eta > 0$ is the learning rate of the inner optimization problem.

(2) \textbf{Update $\omega$}: After obtaining $\theta^{(k+1)}(\omega)$, we use this updated network to update the $\omega$ of DDN. During this process, the loss function depends solely on the parameter $\omega$, allowing $\omega$ to be updated directly by computing its gradient, as shown in the following equation:
\begin{equation}
    \omega^{(k+1)} = \omega^k - \gamma \cdot  \partial_\omega \mathcal{L}\bigl(f_{\theta^{k+1}(\omega)};\{X_{buf}^{n-1},Y_{buf}^{n-1}\} \cup\mathcal{T}^n\bigr)
    \label{omega}
\end{equation}
where $\gamma > 0$ is the learning rate of the outer optimization problem.

\section{Proof of Theorems}
\subsection{Proof of Theorem~\ref{thm:bi_level_ours}}
As aforementioned, Eqn.~(\ref{eqn:ours bi-level}) implements direct parametric optimization by treating the entire memory buffer as learnable parameters. Then we claim that this bi-level optimization framework in Eqn.~(\ref{eqn:ours bi-level}) is equivalent to a gradient matching between the gradients w.r.t. $\theta$ on $\mathcal M^{n-1} \cup \mathcal T^n$ and the entirely parameterized current buffer $\mathcal M^{n}$ in Theorem~\ref{thm:bi_level_ours}. Specifically, the proof of this theorem is presented as follows.

\begin{proof}
Similar to Eqn.~(\ref{theta}), we update the classifier network parameter by one-step SGD at iteration step $k$, that is
\begin{equation}
    \theta^k(\mathcal{M}^n_{tmp}) = \theta^{k-1}-\eta \cdot \nabla_\theta \mathcal{L}(\theta^{k-1}; \mathcal{M}^n_{tmp}).
\label{eq:proof1_1}
\end{equation}
where $\eta > 0$ is the learning rate of the inner optimization problem.

Then we expand the loss function $\mathcal{L}(f_{\theta^n(\mathcal{M}^n_{tmp})};\mathcal{M}^{n-1} \cup\mathcal{T}^n)$ of the outer optimization problem by first-order Taylor expansion around $\theta^{k-1}$, \textit{i.e.},
\begin{equation}
\begin{aligned}
    & \mathcal{L}(f_{\theta^k(\mathcal{M}^n_{tmp})}; \mathcal{M}^{n-1} \cup\mathcal{T}^n)
    \approx \mathcal{L}(f_{\theta^{k-1}}; \mathcal{M}^{n-1} \cup\mathcal{T}^n) \\
    + & \left(\theta^k(\mathcal{M}^n_{tmp})-\theta^{k-1}\right)
    \nabla_\theta \mathcal{L}(\theta^{k-1}; \mathcal{M}^{n-1} \cup \mathcal{T}^n).
\label{eq:proof1_2}
\end{aligned}
\end{equation}
According to Eqn.~(\ref{eq:proof1_1}), we have $\theta^k(\mathcal{M}^n_{tmp})-\theta^{k-1}=-\eta \cdot \nabla_\theta \mathcal{L}(\theta^{k-1}; \mathcal{M}^n_{tmp})$. Substitute it into Eqn.~(\ref{eq:proof1_2}), we obtain
\begin{equation}
\begin{aligned} 
    & \mathcal{L}(f_{\theta^k(\mathcal{M}^n_{tmp})}; \mathcal{M}^{n-1} \cup\mathcal{T}^n) = \mathcal{L}(f_{\theta^{k-1}};\mathcal{M}^{n-1} \cup\mathcal{T}^n) \\
    - & \eta \nabla_\theta \mathcal{L}(\theta^{k-1}; \mathcal{M}^n_{tmp})
    \nabla_\theta \mathcal{L}(\theta^{k-1}; \mathcal{M}^{n-1} \cup \mathcal{T}^n).
\end{aligned}
\label{eq:proof1_3}
\end{equation}
Since the first term in the right-hand-side of Eqn.~(\ref{eq:proof1_3}) is irrelevant to the optimization of variable $\mathcal{M}^n$, it can be ignored in optimization. Therefore, we have the following equivalent optimization problem for the outer loop of Eqn.~(\ref{eqn:ours bi-level}):
\begin{equation}
\begin{aligned} 
          &\min_{\mathcal{M}^n}\mathcal{L}(f_{\theta^k(\mathcal{M}^n_{tmp})};\mathcal{M}^{n-1} \cup\mathcal{T}^n)\Leftrightarrow \\
        & \min_{\mathcal{M}^n}-\langle \nabla_\theta \mathcal{L}(\theta^{k-1}; \mathcal{M}^n_{tmp}),\nabla_\theta \mathcal{L}(\theta^{k-1}; \mathcal{M}^{n-1} \cup \mathcal{T}^n) \rangle.
    \label{theorem}
\end{aligned}
\end{equation}
Therefore, the optimization of Eqn.~(\ref{eqn:ours bi-level}) is equivalent to: 
\begin{equation}
\begin{aligned} 
         \underset{\mathcal{M}^n}{\min}-\langle \nabla_\theta \mathcal{L}(\theta; \mathcal{M}^n),\nabla_\theta \mathcal{L}(\theta; \mathcal{M}^{n-1} \cup \mathcal{T}^n) \rangle .
    \label{theorem}
\end{aligned}
\end{equation}
\end{proof}

\subsection{Proof of Theorem \ref{thm:bi_level_ours2}}
In Sec.~\ref{subsec:ddn}, we propose a hypernetwork-driven parameterization that generates learnable soft labels for all memory buffer samples in Eqn.~(\ref{eqn:ddn}). Furthermore, we introduce Theorem \ref{thm:bi_level_ours2}, accounting that the proposed distillation framework Eqn.~(\ref{eqn:ddn}) is equivalent to a gradient matching between the gradients w.r.t. $\theta$ on $\mathcal M^{n-1} \cup \mathcal T^n$ and the current buffer $\mathcal M^{n}$, which only parameterize a hypernetwork $\mathcal{G}_\omega$. The proof of this theorem is presented as follows.
\begin{proof}
    As shown in Eqn.~(\ref{theta}), we update the classifier network parameter $\theta$ by one-step SGD at iteration step $k$ as follows,
    \begin{equation}
        \theta^k(\omega) = \theta^{k-1} - \eta \cdot \nabla_\theta \mathcal{L}(\theta^{k-1}; \{{X_{buf}^n, \mathcal G_{\omega}^n(X_{buf}^n)}\}).
        \label{eq:proof2_1}
    \end{equation}
    Then we expand the outer optimization loss function by first-order Taylor expansion around $\theta^{k-1}$, that is
    \begin{equation}
    \begin{aligned} 
        &\hspace{6mm} \mathcal{L}(f_{\theta^n(\omega)};\mathcal{M}^{n-1} \cup\mathcal{T}^n)\approx \mathcal{L}(f_{\theta^{k-1}};\mathcal{M}^{n-1} \cup\mathcal{T}^n)  \\
        &+(\theta^k(\omega)-\theta^{n-1})\nabla_\theta \mathcal{L}(\theta^{k-1}; \mathcal{M}^{n-1} \cup \mathcal{T}^n).
    \label{4}
    \end{aligned}
    \end{equation}
    According to Eqn.~(\ref{eq:proof2_1}), we can obtain
    \begin{equation}
    \begin{small}
    \begin{aligned} 
        &\hspace{3mm} \mathcal{L}(f_{\theta^k(\omega)};\mathcal{M}^{n-1} \cup\mathcal{T}^n)\approx \mathcal{L}(f_{\theta^{k-1}};\mathcal{M}^{n-1} \cup\mathcal{T}^n) \\
        &-\eta \cdot
        \nabla_\theta \mathcal{L}(\theta^{n-1}; \{{X_{buf}^n, \mathcal G_{\omega}^n(X_{buf}^n)}\})\nabla_\theta \mathcal{L}(\theta^{n-1}; \mathcal{M}^{n-1} \cup \mathcal{T}^n).
    \label{eq:proof2_2}
    \end{aligned}
    \end{small}
    \end{equation}
    Similarly to the proof of Theorem~\ref{thm:bi_level_ours}, the first term in the right-hand-side of Eqn.~(\ref{eq:proof2_2}) is irrelevant to the optimization of variable $\omega$, which can be ignored in optimization. Thus we have the following equivalent optimization problem:
    \begin{equation}
    \begin{aligned} 
          &\underset{\omega}{\min} \mathcal{L}(f_{\theta^k(\omega)};\mathcal{M}^{n-1} \cup\mathcal{T}^n)\Leftrightarrow\\
         &\underset{\omega}{\min}-\langle \nabla_\theta \mathcal{L}(\theta^{k-1}; \mathcal{M}^n), \nabla_\theta \mathcal{L}(\theta^{k-1}; \mathcal{M}^{n-1} \cup \mathcal{T}^n) \rangle .
    \label{theorem}
    \end{aligned}
    \end{equation}
    Therfore, the optimization of Eqn.~(\ref{eqn:ddn}) is equivalent to:
    \begin{equation}
    \begin{aligned} 
        \min_{\omega}-\bigl\langle \nabla_\theta \mathcal{L}(\theta; \mathcal M^n),\nabla_\theta \mathcal{L}(\theta; \mathcal{M}^{n-1} \cup \mathcal{T}^n) \bigr\rangle .
    \end{aligned}
    \end{equation}
    \end{proof}

\section{Details of Experiments.}

\subsection{Datasets Details}

\textbf{Split CIFAR-10}: Specifically, CIFAR-10 dataset consists of 50,000 training images and 10,000 test images, with each image being a 32×32 RGB image distributed across 10 classes. We partitioned the CIFAR-10 dataset into 5 balanced tasks, each containing 2 classes, referred to as Split CIFAR-10.

\textbf{Split CIFAR-100}: The CIFAR-100 dataset also consists of 50,000 training images and 10,000 test images, with 100 classes in total. We divided it into 10 balanced tasks, each containing 10 classes, denoted as Split CIFAR-100.

\textbf{Split Tiny-ImageNet}: Tiny-ImageNet dataset includes 100,000 training images distributed across 200 classes. This dataset was partitioned into 10 tasks, each with 20 classes, and its test set contains 10,000 images, referred to as Split Tiny-ImageNet.

\subsection{Training Details}
As we aforementioned in Sec.~\ref{subsec:exp_setting}, the primary settings of our experiments are follow DER++~\cite{derpp}. Specifically, we fix the batch size to 32 for both new samples from the current task and previous samples from the memory buffer. We employ the SGD optimizer with a learning rate of 0.03 for training. We train CL models for 1 epoch under online CL, and 50 epochs under offline CL. Additionally, for our DDN, we use the Adam optimizer with learning rates of 0.001, 0.01, and 0.0001 for the Split CIFAR-10, CIFAR-100, and Tiny-ImageNet datasets, respectively. This configuration ensures consistency and reproducibility across our experiments.

\begin{figure}
    \centering
    \subfigure[]{
        \includegraphics[width=0.46\linewidth]{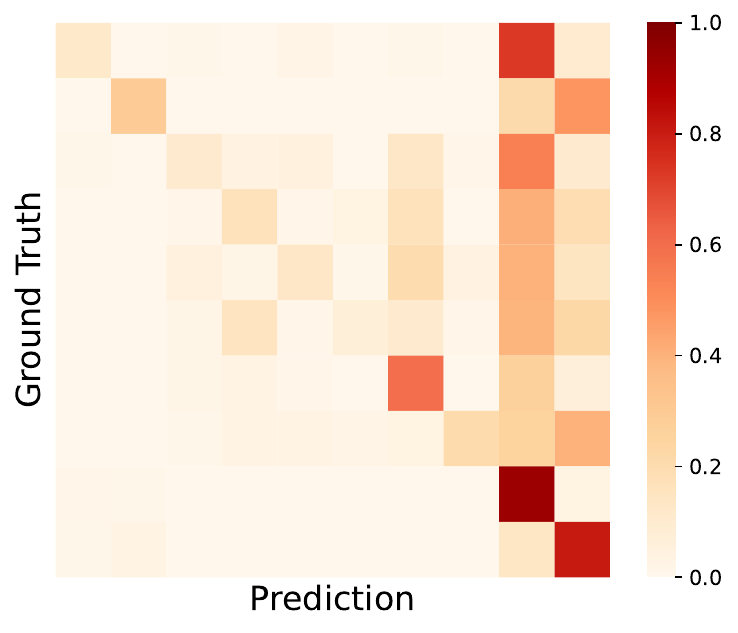}
        \label{subfig:er 200}
    }
    \subfigure[]{
        \includegraphics[width=0.46\linewidth]{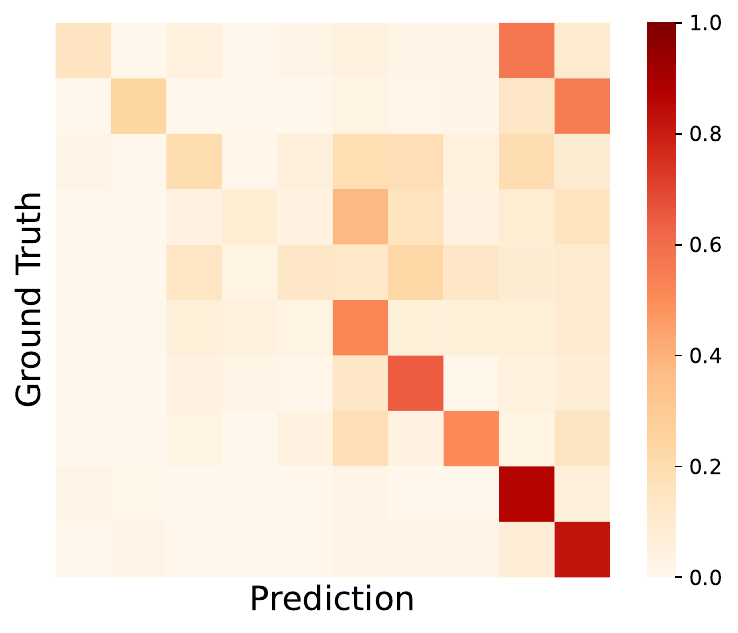}
        \label{subfig:er ddn 200}
    }
    \caption{The normalized confusion matrix of ER and ER DDN based on Split CIFAR-10 with buffer size M=0.2K.}
    \label{fig:confusion}
\end{figure}

\begin{table*}[!htp]
\caption{Comparison for different methods of each task accuracy $a_{t,T}(t=1,2,...T)$ after the whole training based on Split CIFAR-10 with buffer size M=0.2K and M=0.5K.}
\label{each task}
\begin{center}
\centering
    \begin{tabular}{ccccccccc}
\hline
  & method & $a_{1,5}$ & $a_{2,5}$ & $a_{3,5}$ & $a_{4,5}$ & $a_{5,5}$ & ACC$\uparrow$ & FM$\downarrow$ \\ \hline
 & ER & 18.42 & 18.64 & 23.38 & 36.36 & 83.44 & 36.05 & 49.99 \\
 & \cellcolor[HTML]{D8D8D8}ER DDN & \cellcolor[HTML]{D8D8D8}32.45 & \cellcolor[HTML]{D8D8D8}19.95 & \cellcolor[HTML]{D8D8D8}30.47 & \cellcolor[HTML]{D8D8D8}47.12 & \cellcolor[HTML]{D8D8D8}79.71 & \cellcolor[HTML]{D8D8D8}41.94 & \cellcolor[HTML]{D8D8D8}35.29 \\ \cline{2-9}
 & DER++ & 31.27 & 26.12 & 28.63 & 44.27 & 77.42 & 41.54 & 39.50 \\
 & \cellcolor[HTML]{D8D8D8}DER++ DDN & \cellcolor[HTML]{D8D8D8}24.63 & \cellcolor[HTML]{D8D8D8}27.60 & \cellcolor[HTML]{D8D8D8}34.96 & \cellcolor[HTML]{D8D8D8}44.72 & \cellcolor[HTML]{D8D8D8}82.22 & \cellcolor[HTML]{D8D8D8}42.82 & \cellcolor[HTML]{D8D8D8}37.26 \\ \cline{2-9}
 & CLSER & 21.66 & 15.60 & 27.61 & 44.13 & 86.14 & 39.03 & 48.50 \\
 & \cellcolor[HTML]{D8D8D8}CLSER DDN & \cellcolor[HTML]{D8D8D8}27.50 & \cellcolor[HTML]{D8D8D8}19.11 & \cellcolor[HTML]{D8D8D8}31.68 & \cellcolor[HTML]{D8D8D8}45.83 & \cellcolor[HTML]{D8D8D8}80.68 & \cellcolor[HTML]{D8D8D8}40.96 & \cellcolor[HTML]{D8D8D8}40.07 \\ \cline{2-9}
 & ER-ACE & 48.93 & 33.20 & 38.52 & 53.41 & 40.02 & 42.81 & 17.33 \\
\multirow{-9}{*}{M=0.2K} & \cellcolor[HTML]{D8D8D8}ER-ACE DDN & \cellcolor[HTML]{D8D8D8}46.12 & \cellcolor[HTML]{D8D8D8}29.50 & \cellcolor[HTML]{D8D8D8}38.44 & \cellcolor[HTML]{D8D8D8}56.70 & \cellcolor[HTML]{D8D8D8}50.41 & \cellcolor[HTML]{D8D8D8}44.23 & \cellcolor[HTML]{D8D8D8}21.59 \\ \hline
 & ER & 30.35 & 26.03 & 37.51 & 48.02 & 79.86 & 44.35 & 36.90 \\
 & \cellcolor[HTML]{D8D8D8}ER DDN & \cellcolor[HTML]{D8D8D8}40.25 & \cellcolor[HTML]{D8D8D8}34.51 & \cellcolor[HTML]{D8D8D8}39.54 & \cellcolor[HTML]{D8D8D8}49.03 & \cellcolor[HTML]{D8D8D8}77.25 & \cellcolor[HTML]{D8D8D8}48.11 & \cellcolor[HTML]{D8D8D8}32.30 \\ \cline{2-9}
 & DER++ & 39.03 & 36.97 & 36.63 & 50.51 & 82.44 & 49.11 & 34.24 \\
 & \cellcolor[HTML]{D8D8D8}DER++ DDN & \cellcolor[HTML]{D8D8D8}40.70 & \cellcolor[HTML]{D8D8D8}36.26 & \cellcolor[HTML]{D8D8D8}38.23 & \cellcolor[HTML]{D8D8D8}54.43 & \cellcolor[HTML]{D8D8D8}77.82 & \cellcolor[HTML]{D8D8D8}49.48 & \cellcolor[HTML]{D8D8D8}28.19 \\ \cline{2-9}
 & CLSER & 34.10 & 27.13 & 32.63 & 47.99 & 81.09 & 44.58 & 37.31 \\
 & \cellcolor[HTML]{D8D8D8}CLSER DDN & \cellcolor[HTML]{D8D8D8}40.74 & \cellcolor[HTML]{D8D8D8}34.37 & \cellcolor[HTML]{D8D8D8}32.71 & \cellcolor[HTML]{D8D8D8}53.73 & \cellcolor[HTML]{D8D8D8}75.29 & \cellcolor[HTML]{D8D8D8}47.37 & \cellcolor[HTML]{D8D8D8}29.75 \\ \cline{2-9}
 & ER-ACE & 52.88 & 32.42 & 49.68 & 58.02 & 42.13 & 47.02 & 17.54 \\
\multirow{-8}{*}{M=0.5K} & \cellcolor[HTML]{D8D8D8}ER-ACE DDN & \cellcolor[HTML]{D8D8D8}52.31 & \cellcolor[HTML]{D8D8D8}34.20 & \cellcolor[HTML]{D8D8D8}43.28 & \cellcolor[HTML]{D8D8D8}62.11 & \cellcolor[HTML]{D8D8D8}54.68 & \cellcolor[HTML]{D8D8D8}49.31 & \cellcolor[HTML]{D8D8D8}16.84 \\ \hline
\end{tabular}
\end{center}

\end{table*}

\subsection{Comparison methods}
We provide a detailed induction of the four baselines and other comparison methods mentioned in the Table \ref{tab:online_comparison}:

\textbf{ER}~\cite{ERratcliff1990connectionist} is the most common replay-based continual learning method, 
which trains the new samples with the memory buffer samples together. Specifically, under the online CL setting, ER employs reservoir sampling to construct the memory buffer, which can be regarded as a balanced dataset to approximate the data distribution of all encountered tasks.

\textbf{DER++}~\cite{derpp}
builds upon ER by integrating an additional knowledge distillation, which calculates the MSE loss between the logits predicted by the current model and previous logits predicted by the old model. To reduce the memory size, DER++ directly stores previous logits with the memory buffer samples.

\textbf{CLSER}~\cite{clser}
incorporates a fast learning module for adapting to new tasks and a slow learning module for stabilizing old knowledge integration based on a complementary learning system (CLS). The collaboration between these two modules significantly reduce forgetting in CL.

\textbf{ER-ACE}\cite{ERACE} shields old classes during updates and uses asymmetric update rules to encourage new classes to adapt to old ones.
For the incoming new samples, ER-ACE masks the predicted probabilities corresponding to old classes, thereby preventing the gradients of new samples from negatively impacting the discriminative ability of the old classes.

\textbf{iCarl}~\cite{icarl} proposes the nearest class mean (NCM) classification rule, selecting old training samples closest to the feature mean of each class for replay. It leverages knowledge distillation and prototype rehearsal for representation learning.

\textbf{LUCIR}~\cite{LUCIR}
aims to balance the influence of new and old tasks. Specifically, ER-ACE employs cosine normalization to address the imbalance in embeddings between new and old classes. It further aligns embeddings provided by new and old models by minimizing their discrepancy. Additionally, it also maximizes the inter-class differences to enhance the model's discriminative ability.

\textbf{BIC}~\cite{BIC} 
introduces an affine transformation to the classifier to correct the classification bias in the CL process. This method adopts a two-stage training approach. In the first stage, the classification model is trained by ER, while in the second stage, the affine transformation is trained by the memory buffer, which is an approximation of all data distributions.

\textbf{LoDM}~\cite{LoDM} proposed a plugin module for previous dataset distillation methods. For example, based on DM~\cite{DMzhao2023dataset}, LoDM decomposed the synthetic images as two low-rank matrices to reduce the optimization parameters in image space, which significantly reduces the calculation in bi-level optimization. However, LoDM treats CL solely as a downstream application and performs distillation for each individual task, which easily neglects the global relationship between sequential tasks.

\textbf{Mnemonics}~\cite{liu2020mnemonics} optimized examples saved in the memory buffer for each task individually as we introduced in Eqn.~(\ref{eqn:Datadistillation}). This formulation only distilled the intra-task information for the current task while ignoring the correlation between different tasks. 

\begin{figure*}[!t]
    \centering
    \subfigure[]{
        \includegraphics[width=0.18\linewidth]{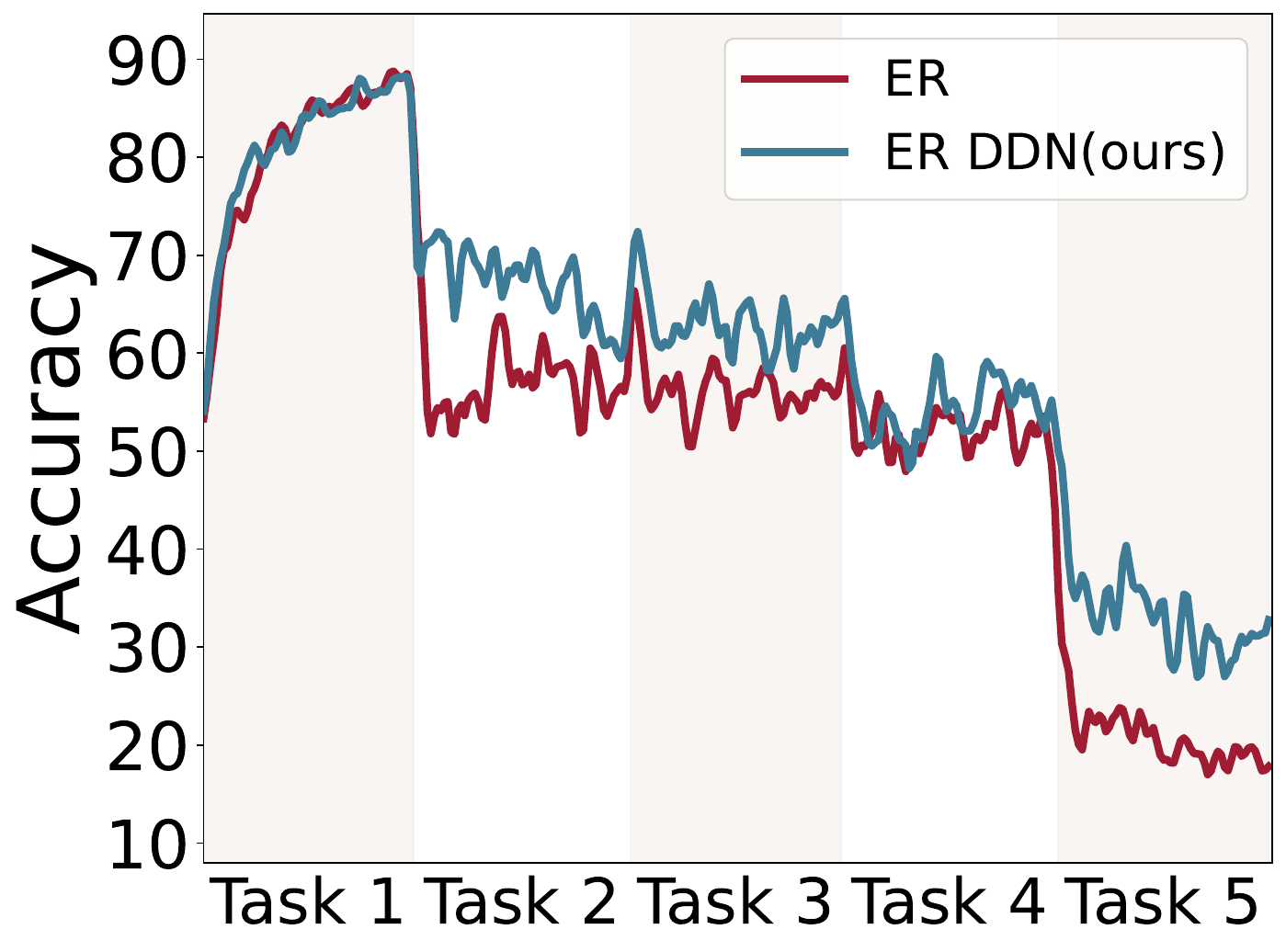}
        \label{subfig:label 0}
    }
    \subfigure[]{
        \includegraphics[width=0.18\linewidth]{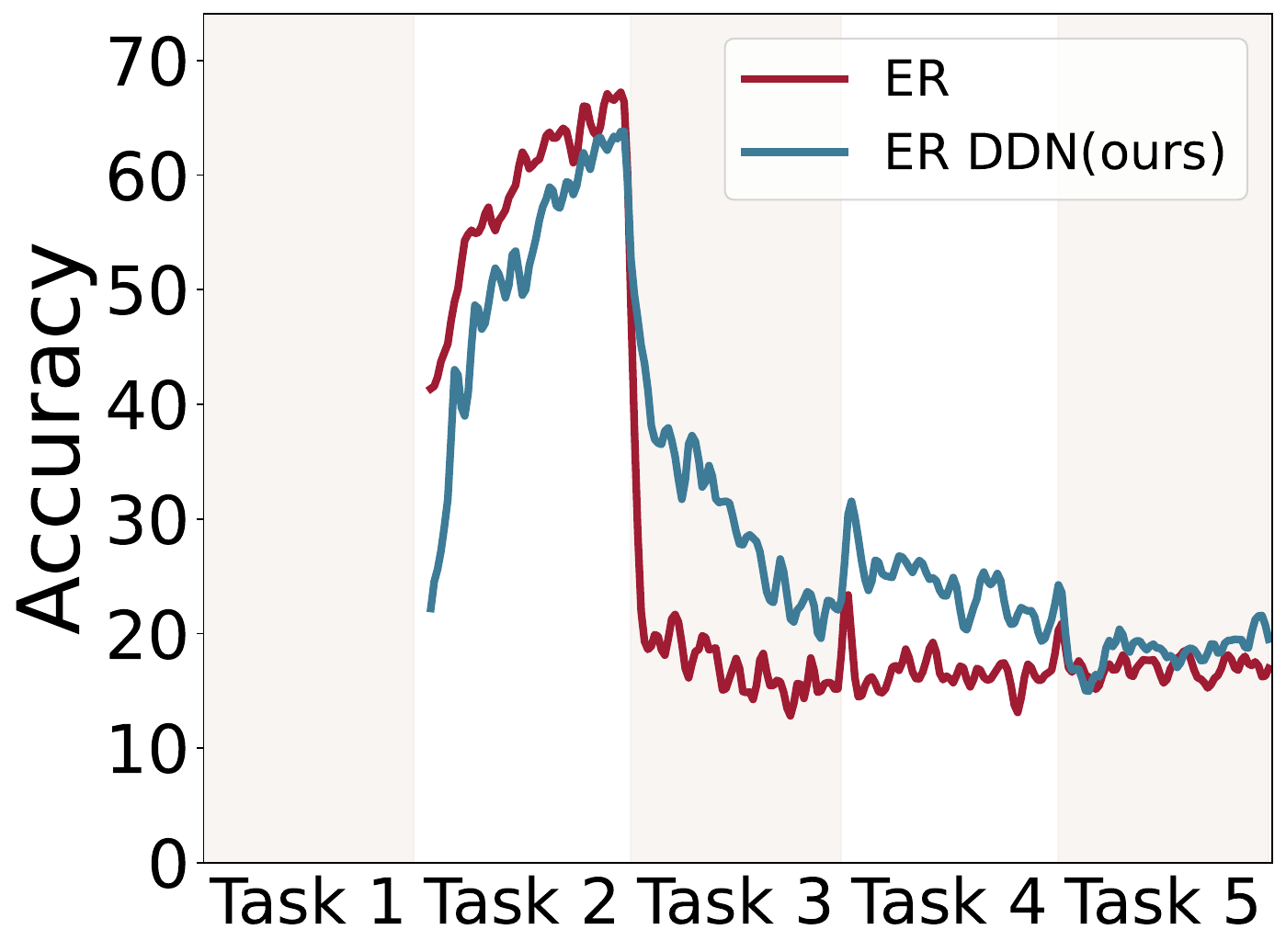}
        \label{subfig:label 1}
    }
    \subfigure[]{
        \includegraphics[width=0.18\linewidth]{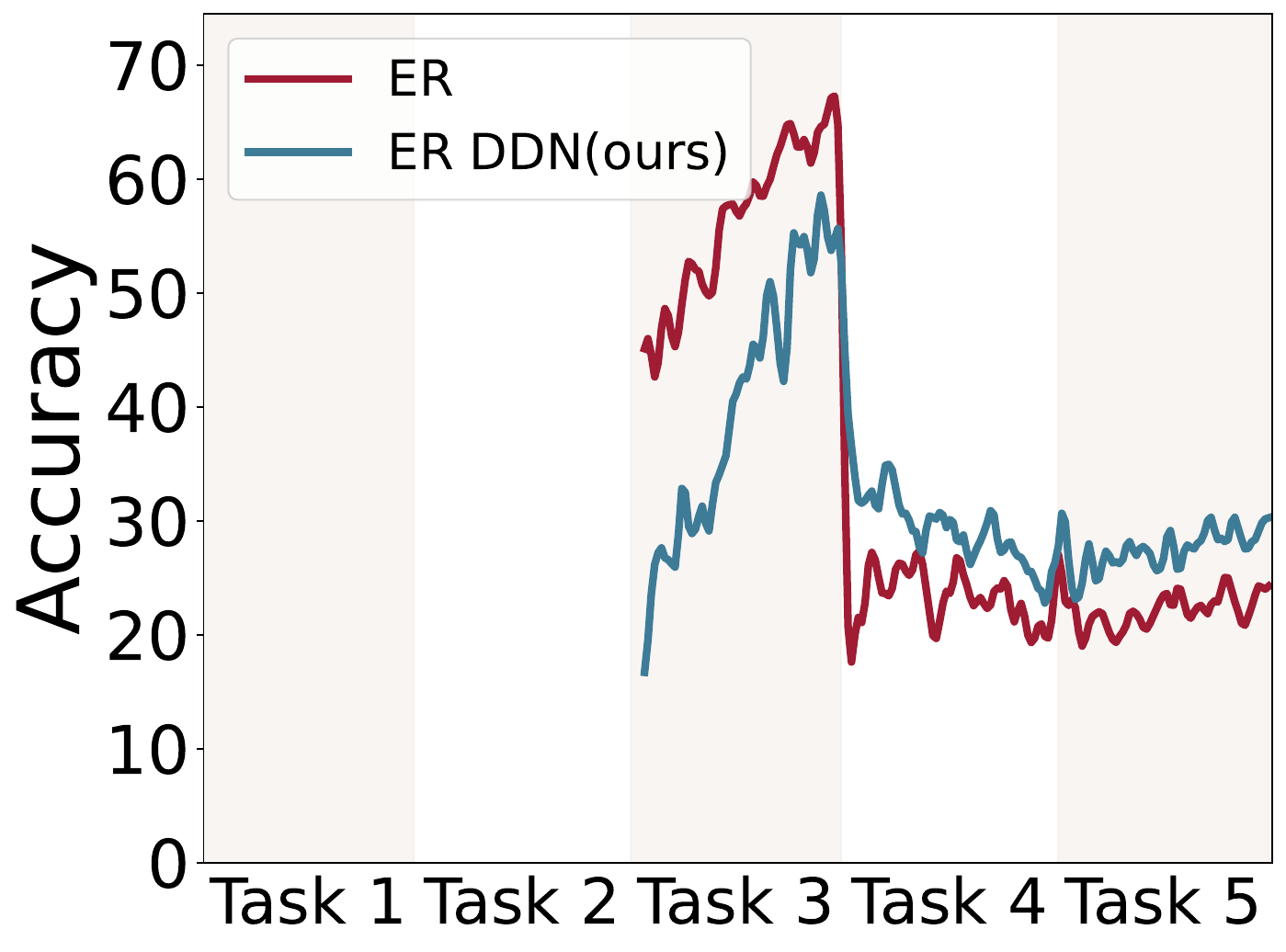}
        \label{subfig:label 2}
    }
    \subfigure[]{
        \includegraphics[width=0.18\linewidth]{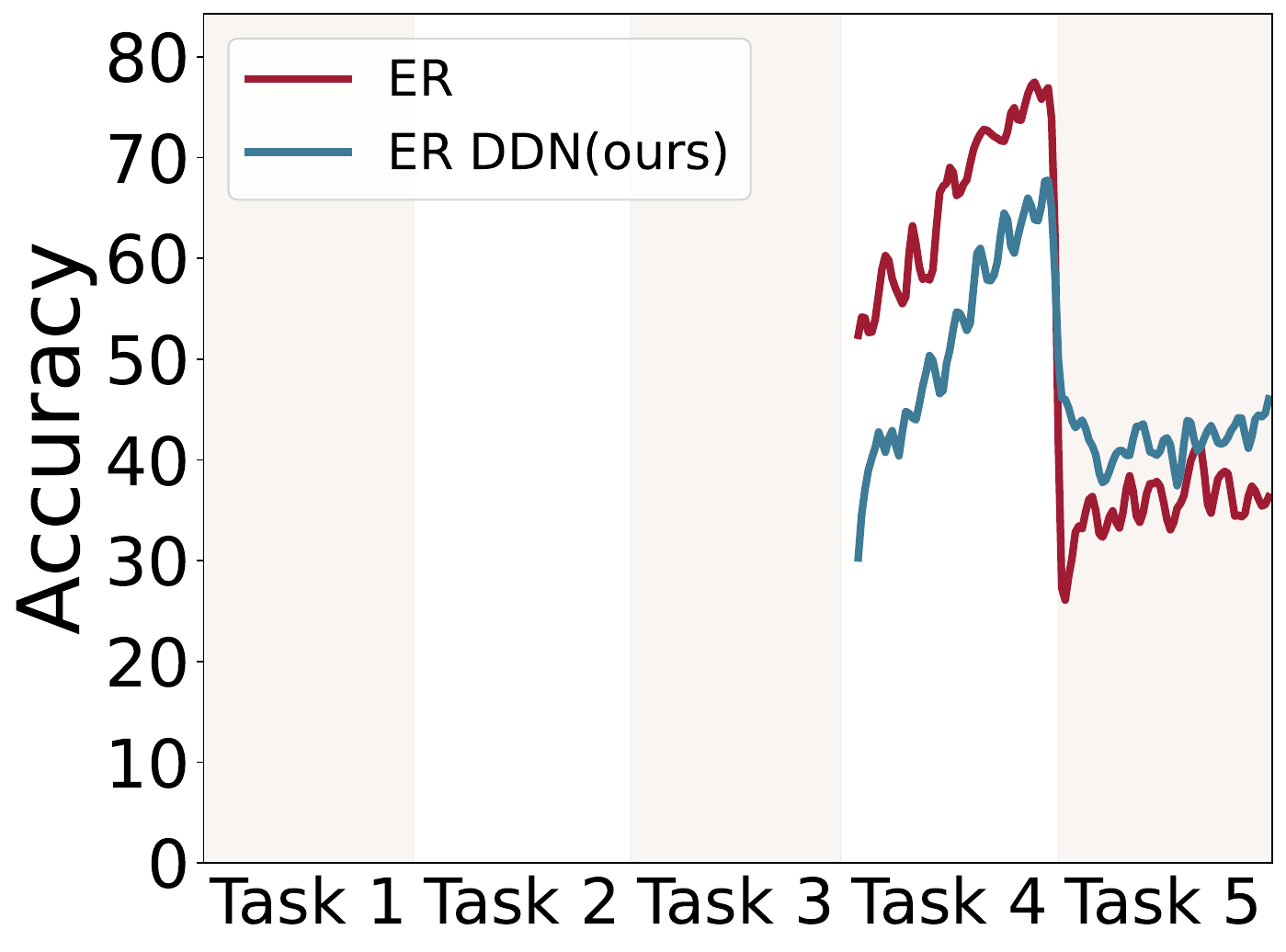}
        \label{subfig:label 3}
    }
    \subfigure[]{
        \includegraphics[width=0.18\linewidth]{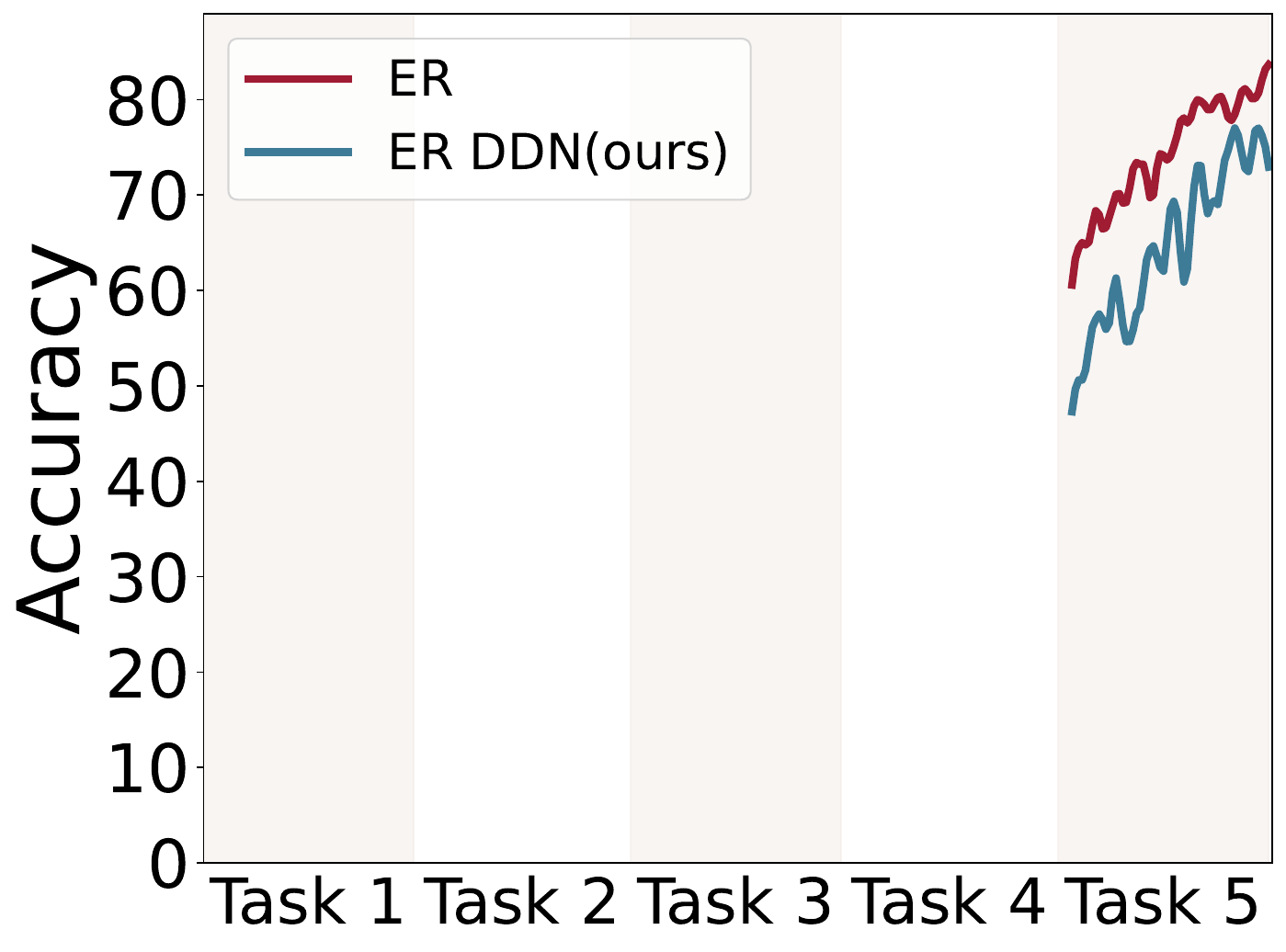}
        \label{subfig:label 4}
    }
    \caption{The accuracy of each task in the whole training process on Split CIFAR-10 with buffer size M=0.2K .}
    \label{fig:acc_auc}
    \vskip -0.1in
\end{figure*}

\section{Addational experiments}

\subsection{Confusion matrix}
We visualized the confusion matrices of the baseline ER and our proposed ER DDN to better illustrate their performance in classifying samples from different classes. The horizontal axis of the confusion matrix represents the predicted classes, while the vertical axis represents the ground truth classes. Values along the diagonal indicate the number of correctly classified samples. As shown in the Fig.~\ref{fig:confusion}, Fig.~\ref{subfig:er 200} represents the confusion matrix for ER. It is evident that ER exhibits a strong bias towards predicting test samples as belonging to new classes. This bias leads to a higher misclassification rate for samples from old classes, exacerbating catastrophic forgetting. In contrast, due to the incorporation of soft-label distillation, ER DDN(ours) in Fig.~\ref{subfig:er ddn 200}, demonstrates superior retention of old-class information and effectively mitigates the bias towards new classes. During testing, predictions are more balanced between old and new classes, resulting in an overall improvement in classification performance.\\

\subsection{Accuracy of each task}
To comprehensively analyze how our method alleviates forgetting, we evaluated the accuracy of all tasks at the end of training, $i.e.$, $a_{t,T} (t=1,2,...T)$  to assess the model's performance across different tasks. Under the online learning setting, we compared our method with various baselines on the Split CIFAR-10 dataset with buffer sizes of M=0.2K and M=0.5K.\\
The experimental results demonstrate that our method significantly improves the classification accuracy of old classes, particularly for ER with buffer sizes of M=0.2K and M=0.5K, the improvement in old-class accuracy is especially notable. This enhancement effectively mitigates the bias toward new classes, thereby improving overall performance. Further analysis reveals that the performance gains primarily stem from better accuracy on old classes. Specifically, when the buffer size is M=0.2K, our method reduces the forgetting measure (FM) by 14.7\%, highlighting its strong capability in alleviating forgetting and preserving old-class information.

\subsection{Analyse of each task accuracy} 
To investigate the accuracy trends of each task throughout the training process, Fig.~\ref{fig:acc_auc} illustrates a comparison between the baseline ER and our proposed ER DDN on the Split CIFAR-10 dataset with buffer size M=0.2K. Each subplot represents the accuracy progression of task $i$ over the course of training. The results clearly show that our method consistently outperforms ER across the entire process. Notably, after a task transitions to being an old task, its accuracy during subsequent training remains significantly higher than that of ER. This indicates that our approach is more effective at retaining knowledge from old tasks, significantly mitigating catastrophic forgetting. These findings further validate the effectiveness of our method DDN in consolidating old task knowledge within the continual learning framework.



\end{document}